\documentclass[10pt,journal,cspaper,compsoc]{IEEEtran}
\usepackage{amsmath}
\usepackage{amsfonts}
\usepackage{mathrsfs}
\usepackage{amssymb}
\usepackage{color}

\usepackage[compress]{cite}
\usepackage{graphicx}
\usepackage{epstopdf}

\usepackage{url}
\usepackage{algorithm}
\usepackage{algorithmic}
\usepackage{subfig}
\usepackage{array}
\usepackage[
            bookmarksnumbered,
            bookmarksopen,
            colorlinks,
            linkcolor=black,
            anchorcolor=black,
            citecolor=black
            ]{hyperref}

\newtheorem{thm}{Theorem}
\newtheorem{lema}{Lemma}
\newtheorem{rema}{Remark}
\newtheorem{prop}{Proposition}

\ifCLASSOPTIONcompsoc
\else
\fi
\ifCLASSINFOpdf
\else
\fi
\begin{document}
%
\title{Classification with Noisy Labels\\ by Importance Reweighting}
%
%
%
%

\author{Tongliang~Liu~
        and~Dacheng~Tao,~\IEEEmembership{Fellow,~IEEE}

\IEEEcompsocitemizethanks{\IEEEcompsocthanksitem T. Liu and D. Tao are with the Centre for Quantum Computation $\&$ Intelligent Systems and the Faculty of Engineering and Information Technology, University of Technology, Sydney, 81 Broadway Street, Broadway, NSW 2007, Australia.\protect\\
E-mail: tliang.liu@gmail.com, dacheng.tao@uts.edu.au.
}
\thanks{}}

%
%

\markboth{IEEE Transactions on Pattern Analysis and Machine Intelligence}
{Shell \MakeLowercase{\textit{et al.}}: Bare Demo of IEEEtran.cls for Computer Society Journals}
%


\IEEEcompsoctitleabstractindextext{%
\begin{abstract}
In this paper, we study a classification problem in which sample labels are randomly corrupted. In this scenario, there is an unobservable sample with noise-free labels. However, before being observed, the true labels are independently flipped with a probability $\rho\in[0,0.5)$, and the random label noise can be class-conditional. Here, we address two fundamental problems raised by this scenario. The first is how to best use the abundant surrogate loss functions designed for the traditional classification problem when there is label noise. We prove that any surrogate loss function can be used for classification with noisy labels by using importance reweighting, with consistency assurance that the label noise does not ultimately hinder the search for the optimal classifier of the noise-free sample. The other is the open problem of how to obtain the noise rate $\rho$. We show that the rate is upper bounded by the conditional probability $P(\hat{Y}|X)$ of the noisy sample. Consequently, the rate can be estimated, because the upper bound can be easily reached in classification problems. Experimental results on synthetic and real datasets confirm the efficiency of our methods.
\end{abstract}

}

\maketitle

\IEEEdisplaynotcompsoctitleabstractindextext

%
\IEEEpeerreviewmaketitle

\section{Introduction}
%
%

%
%
%
%
\IEEEPARstart{C}{lassification} crucially relies on the accuracy of the dataset labels. In some situations, observation labels are easily corrupted and, therefore, inaccurate. Designing learning algorithms that account for noisy labeled data is therefore of great practical importance and has attracted a significant amount of interest in the machine learning community.

The random classification noise (RCN), in which each label is flipped independently with a probability $\rho\in[0,0.5)$, has been proposed; it was proven to be PAC-learnable by Angluin and Laird \cite{angluin1988learning} soon after the noise-free PAC learning model was introduced by Valiant \cite{valiant1984theory}. Many related works then followed: Kearns \cite{kearns1998efficient} proposed the statistical query model to learn with RCN. The restriction he enforced is that learning is based not on the particular properties of individual random examples, but instead on the global statistical properties of large samples. Such an approach to learning seems intuitively more robust. Lawrence and Scholk{\"o}pf \cite{lawrence2001estimating} proposed a Bayesian model for this noise and applied it to sky moving in images. Biggio et al. \cite{journals/jmlr/BiggioNL11} enabled support vector machine learning with RCN via a kernel matrix correction. And Yang et al. \cite{yang2012multiple} developed multiple kernel learning for classification with noisy labels using stochastic programming. The interested reader is referred to further examples in the survey \cite{6685834}. However, most of these algorithms are designed for specific surrogate loss functions, and the use and benefit of the large number of surrogate loss functions designed for the traditional (noise-free) classification problem is important to investigate in order to solve classification problems in the presence of label noise.


Aslam and Decatur \cite{aslam1996sample} proved that the RCN exploited using a 0-1 loss function is PAC-learnable if the function class is of finite VC-dimension. Manwani and Sastry \cite{manwani2013noise} analyzed the tolerance properties of RCN for risk minimization under several frequently used surrogate loss functions and showed that many of them do not tolerate RCN. Natarajan et al. \cite{natarajan2013learning} reported two methods for learning asymmetric RCN models, in which the random label noise is class-conditional. Their methods exploit many different surrogate loss functions \cite{bartlett2006convexity}: the first model uses unbiased estimators of surrogate loss functions for empirical risk minimization, but the unbiased estimator may be non-convex, even if the original surrogate loss function is convex; their second method uses label-dependent costs. The latter approach is based on the idea that there exists an $\alpha\in(0,1)$ such that the minimizer of the expected risk as assessed using the $\alpha$-weighted 0-1 loss function $\ell_\alpha(t,y)=(1-\alpha)1_{y=1}1_{t\leq 0}+\alpha 1_{y=-1}1_{t>0}$ over the noisy sample distribution, where $t$ is the predicted value and $y$ is the label of the example, has the same sign as that of the Bayes classifier which minimizes the expected risk as assessed using the 0-1 loss function over the clean sample distribution; see, for example, Theorem 9 in \cite{natarajan2013learning}. The method is notable because it can be applied to all the convex and classification-calibrated surrogate loss functions (If a surrogate loss function is classification-calibrated and the sample size is sufficiently large, the surrogate loss function will help learn the same optimal classifier as the 0-1 loss function does, see Theorem 1 in Bartlett et al. \cite{bartlett2006convexity}). This modification is based on the asymmetric classification-calibrated results \cite{scott2012calibrated} and cannot be used to improve the performance of symmetric RCN problems or the algorithms that employ the non-classification-calibrated surrogate loss functions.

To best use and benefit from the abundant surrogate loss functions designed for the traditional classification problems, here we propose an importance reweighting method in which any surrogate loss function designed for a traditional classification problem can be used for classification with noisy labels. In our method, the weights are non-negative, so the convexity of objective functions does not change. In addition, our method inherits most batch learning optimization procedures designed for traditional classification problems with different regularizations; see, for examples, \cite{belkin2006manifold,zou2005regularization,pong2010trace,nesterov2005smooth,gong2013general}.

Although many works have focused on the RCN model, how to best estimate the noise rate $\rho$ remains an open problem \cite{yang2012multiple} and severely limits the practical application of the existing algorithms. Most previous works make the assumption that the noise rate is known or learn it using cross-validation, which is time-consuming and lacks a guarantee of generalization accuracy. In this paper, we set the noise rate to be asymmetric and unknown and denote the flip probability of positive labels $P(\hat{Y}=-1|Y=+1)$ and the flip probability of negative labels $P(\hat{Y}=+1|Y=-1)$ by $\rho_{+1}$ and $\rho_{-1}$, respectively. We show that the noise rate $\rho_{+1}$ (or $\rho_{-1}$) is upper bounded by the conditional probability $P(-1|X)$ (or $P(+1|X)$) of the noisy data. Moreover, the upper bound can be reached if there exists an $x\in\mathcal{X}$ such that the probability $P(+1|x)$ (or $P(-1|x)$) of the ``clean" sample is zero, which is very likely to hold for classification problems. The noise rates $\rho_{\pm 1}$ are therefore estimated by finding the minimal $P(\mp 1|X)$ of the noisy training sample.

\subsection{Related Works}
Kearns and Li \cite{kearns1993learning} introduced the malicious noise (MN) model, in which an adversary can access the sample and randomly replace a fraction of them with adversarial ones. It has been proven that any nontrivial target function class cannot be PAC learned with accuracy $\epsilon$ and malicious noise rate $\eta\geq\epsilon(1+\epsilon)$; see, for examples, \cite{kearns1993learning,cesa1999sample,bshouty2002pac}. Long and Servedio \cite{long2011learning} proved that an algorithm for learning $\gamma$-margin half-spaces that minimizes a convex surrogate loss function for misclassification risk cannot tolerate malicious noise at a rate greater than $\mathcal{O}(\epsilon\gamma)$. They therefore proposed an algorithm, that does not optimize a convex loss function and that can tolerate a higher rate of malicious noise than order $\mathcal{O}(\epsilon\gamma)$. Further details about the MN model can be found in \cite{klivans2009learning}.

Cesa-Bianchi et al. \cite{cesa2011online} considered a more complicated model in which the features and labels are both added with zero-mean and variance-bounded noise. They used unbiased estimates of the gradient of the surrogate loss function to learn from the noisy sample in an online learning setting. Perceptron algorithms that tolerate RCN have also been widely studied; see, for examples, \cite{bylander1994learning,cohen1997learning,blum1998polynomial,stempfel2007learning}. See Khardon and Wachman \cite{khardon2007noise} for a survey of noise-tolerant variants of perceptron algorithms.

As well as these model-motivated algorithms, many algorithms that exploit robust surrogate loss functions have been designed for learning with any kind of feature and label noise. Robust surrogate loss functions, such as the Cauchy loss function \cite{moore1977robust} and correntropy (also known as the Welsch loss function),  \cite{liu2007correntropy,he2011maximum}, have been empirically proven to be robust to noise. Some other algorithms, such as confidence weighted learning \cite{crammer2010learning}, have also been proposed for noise-tolerant learning.

To the best of our knowledge, the only work that related to learn the unknown noise rate was proposed by Scott et al. \cite{conf/colt/ScottBH13}. Inspired by the theory of mixture proposition estimation \cite{blanchard2010semi}, they provided estimators for the inversed noise rates $\pi_{+1} = P(Y=-1|\hat{Y}=+1)$ and $\pi_{-1} = P(Y=+1|\hat{Y}=-1)$. However, there were no efficient algorithms that can be used to calculate the estimators until Scott \cite{scott2015a} proposed an efficient algorithm for optimizing them during the preparation of this manuscript. By using Bayes' rule, we have $P(\hat{Y}|Y)=P(Y|\hat{Y})P(\hat{Y})/P(Y)$. However, our method for estimation the noise rates is essentially different from that of Scott et al. \cite{conf/colt/ScottBH13} because $P(Y)$ is unknown. The inversed noise rates can be used to design algorithms for classification with label noise; see, for example, \cite{scott2015a}. In this paper, we also design importance reweighting algorithms for classification with label noise by employing the inversed noise rates.

The rest of this paper is organized as follows. The problem is set up in Section \ref{sec2}. Section \ref{sec3} presents some useful results applied to the traditional classification problem. In Section \ref{method1}, we discuss how to perform classification in the presence of RCN and benefit from the abundant surrogate loss functions and algorithms designed for the traditional classification problem. In Section \ref{sec5}, we discuss how to reduce the uncertainty introduced by RCN by estimating the conditional probability $P(\hat{Y}|X)$ of the noisy sample; theoretical guarantees for the consistency of the learned classifiers are provided; certain convergence rates are also characterized in this section. In Section \ref{sec6}, an approach for estimating the noise rates is proposed. We also provide a detailed comparison between the theory of noise rate estimation and that of the inversed noise rate estimation in this section. We present the proofs of our assertions in Section \ref{proofsection}. In Section \ref{exp}, we present experimental results on synthetic and benchmark datasets, before concluding in Section \ref{con}.

\section{Problem Setup}\label{sec2}
Let $D$ be the distribution of a pair of random variables $(X,Y)\in \mathcal{X}\times\{\pm 1\}$, where $\mathcal{X}\subseteq\mathbb{R}^m$. Our goal is to predict a label for any given observation $X\in\mathcal{X}$ using a sample drawn i.i.d. from the distribution $D$. However, in many real-world classification problems, sample labels are randomly corrupted. We therefore consider the asymmetric RCN model (see \cite{natarajan2013learning}). Let $(X_1,Y_1),\ldots,(X_n,Y_n)$ be an i.i.d. sample drawn from the distribution $D$ and $(X_1,\hat{Y}_1),\ldots,(X_n,\hat{Y}_n)$ the corresponding corrupted ones. The asymmetric RCN model is given by:
\begin{eqnarray*}
P(\hat{Y}=+1|Y=-1)=\rho_{-1}, P(\hat{Y}=-1|Y=+1)=\rho_{+1},
\end{eqnarray*}
where $\rho_{+1},\rho_{-1}\in[0,1)$ and $\rho_{+1}+\rho_{-1}< 1$.

We denote by $D_\rho$ the distribution of the corrupted variables $(X,\hat{Y})$. In our setting, the ``clean" sample $(X_1,Y_1),\ldots,(X_n,Y_n)$ and the noise rates $\rho_{+1}$ and $\rho_{-1}$ are not available for learning algorithms. The classifier and noise rates are learned only by using the knowledge from the corrupted sample $(X_1,\hat{Y}_1),\ldots,(X_n,\hat{Y}_n)$.


\section{The Traditional Classification Pr-oblem}\label{sec3}
Classification is a fundamental machine learning problem. One intuitive way to learn the classifier is to find a decision function $f\in F$, such that the expected risk $R_{1,D}(f)=E_{(X,Y)\sim D}[1_{\text{sign}(f(X))\neq Y}]$ is minimized, where $F$ is the function class for searching. However, two problems remain when minimizing the expected risk: first, that the 0-1 loss function is neither convex nor smooth, and second that the distribution $D$ is unknown. The solutions to these two problems are summarized below.

For the problem that the 0-1 loss function is neither convex nor smooth, abundant convex surrogate loss functions (most are smooth) with the classification-calibrated property \cite{bartlett2006convexity,scott2012calibrated} have been proposed. These surrogate loss functions, such as square loss, logistic loss, and hinge loss, have been proven useful in many real-world applications. Apart from the convex classification-calibrated surrogate loss functions, many other non-convex surrogate loss functions empirically proven to be robust to noise, such as Cauchy loss and Welsch loss, are also frequently used. 
In this paper, we show that all these surrogate loss functions, as well as the non-classification-calibrated surrogate loss functions, such as the asymmetric exponential loss function (see Example 8 in\cite{bartlett2006convexity})
\begin{eqnarray*}
\ell(t,y)=
   \begin{cases}
   \exp(-2ty) &t\leq 0\\
   \exp(-ty) &t>0,
   \end{cases}
\end{eqnarray*}
can be used directly for classification in the presence of RCN by employing the importance reweighting method.

For the problem that distribution $D$ is unknown, empirical risk is proposed to approximate the expected risk. The empirical risk is defined as
\begin{eqnarray*}
&&\hat{R}_{\ell,D}(f)=\frac{1}{n}\sum_{i=1}^{n}\ell(f(X_i),Y_i),
\end{eqnarray*}
where the corresponding expected risk is
\begin{eqnarray*}
&&R_{\ell,D}(f)=R[D,f,\ell]=E_{(X,Y)\sim D}[\hat{R}_{\ell,D}(f)]
\end{eqnarray*}
and $\ell$ denotes any surrogate loss function. The classifier is then learned by empirical risk minimization (ERM) \cite{vapnik2000nature}:
\begin{eqnarray*}
&&f_n=\arg\min_{f\in F}\hat{R}_{\ell,D}(f).
\end{eqnarray*}

The consistency of $R_{\ell,D}(f_n)$ to $\min_{f\in F}R_{\ell,D}(f)$ is therefore essential for designing surrogate loss functions and learning algorithms. Let
\begin{eqnarray*}
&&f^*=\arg\min_{f\in F}R_{\ell,D}(f).
\end{eqnarray*}
It \cite{anthony2009neural} is easily proven that
\begin{eqnarray*}
&&R_{\ell,D}(f_n)-R_{\ell,D}(f^*)\leq 2\sup_{f\in F}|R_{\ell,D}(f)-\hat{R}_{\ell,D}(f)|.
\end{eqnarray*}
The right hand side term is known as the generalization error, and the consistency is guaranteed by convergence of the generalization error. We note that learning algorithms which are based on ERM, such as those using Tikhonov or manifold regularization, will not have a slower convergence rate of consistency than that of ERM. In this paper, we therefore provide consistency guarantees for learning algorithms dealing with RCN by deriving the generalization error bounds of the corresponding ERM algorithms.

\section{Learning with Importance Reweigh-ting}\label{method1}
Importance reweighting is widely used for domain adaptation \cite{bruzzone2010domain}, but here we introduce it to classification in the presence of label noise. One observation \cite{gretton2009covariate} from the field of importance reweighting is as follows:
\begin{eqnarray*}
&R_{\ell,D}(f)&=R[D,f,\ell]=E_{(X,Y)\sim D}[\ell(f(X),Y)]\\
&&=E_{(X,\hat{Y})\sim D_\rho}\left[\frac{P_{D}(X,Y)}{P_{D_\rho}(X,\hat{Y})}\ell(f(X),\hat{Y})\right]\\
&&=R\left[D_\rho,f,\frac{P_{D}(X,Y)}{P_{D_\rho}(X,\hat{Y})}\ell(f(X),\hat{Y})\right]\\
&&=R\left[D_\rho,f,\beta(X,\hat{Y})\ell(f(X),\hat{Y})\right]\\
&&=R_{\beta\ell,D_\rho}(f),\\
\end{eqnarray*}
where $\beta(X,\hat{Y})=\frac{P_{D}(X,Y)}{P_{D_\rho}(X,\hat{Y})}$.

For the problem of classification in the presence of label noise, note that $P_D(X)=P_{D_\rho}(X)$. We therefore have
\begin{eqnarray*}
&\beta(X,\hat{Y})&=\frac{P_{D}(X,Y)}{P_{D_\rho}(X,\hat{Y})}=\frac{P_D(Y|X)P_D(X)}{P_{D_\rho}(\hat{Y}|X)P_{D_\rho}(X)}\\
&&=\frac{P_{D}(Y|X)}{P_{D_\rho}(\hat{Y}|X)}.\\
\end{eqnarray*}
Thus, even though the labels are corrupted, classification can still be implemented if only the weight $\beta(X,\hat{Y})=P_{D}(Y|X)/P_{D_\rho}(\hat{Y}|X)$ could be accessed to the loss $\ell(f(X),\hat{Y})$.

\begin{lema}\label{weight}
The asymmetric RCN problem can be addressed by reweighting the surrogate loss functions of the traditional classification problem via importance reweighting. The weight given to a noisy example $(X,\hat{Y})\sim D_\rho$ is
\begin{eqnarray*}
&\beta(X,\hat{Y})&=\frac{P_{D}(Y|X)}{P_{D_\rho}(\hat{Y}|X)}\\
&&=\frac{P_{D_\rho}(\hat{Y}|X)-\rho_{-\hat{Y}}}{(1-\rho_{+1}-\rho_{-1})P_{D_\rho}(\hat{Y}|X)}.
\end{eqnarray*}
The weight $\beta(X,\hat{Y})$ is non-negative if $P_{D_\rho}(\hat{Y}|X)\neq 0$. If $P_{D_\rho}(\hat{Y}|X)= 0$, we intuitively let $\beta(X,\hat{Y})= 0$.
\end{lema}

A classifier can therefore be learned for the ``clean'' data in the presence of asymmetric RCN by minimizing the following reweighted empirical risk:
\begin{eqnarray*}\label{ourmodel}
&\hat{f}_{n}&=\arg\min_{f\in F}\hat{R}_{\beta\ell,D_\rho} \\
&&=\arg\min_{f\in F}\frac{1}{n}\sum_{i=1}^{n}{\beta}(X_i,\hat{Y}_i)\ell(f(X_i),\hat{Y}_i),
\end{eqnarray*}
where
\begin{eqnarray*}
&&{\beta}(X_i,\hat{Y}_i)=\frac{P_{D_\rho}(\hat{Y}_i|X_i)-\rho_{-\hat{Y}_i}}{(1-\rho_{+1}-\rho_{-1})P_{D_\rho}(\hat{Y}_i|X_i)}.
\end{eqnarray*}

By the following proposition, based on Talagrand's Lemma (see, e.g., Lemma 4.2 in \cite{mohri2012foundations}), we show that, given $P_{D_\rho}(\hat{Y}|X)$, the above weighted empirical risk will converge to the unweighted expected risk of the ``clean" data for any $f\in F$. So, $R_{\ell,D}$ can be approximated by $\hat{R}_{\beta\ell,D_\rho}$.
\begin{prop}\label{rademacher}
Given the conditional probability $P_{D_\rho}(\hat{Y}|X)$ and the noise rates $\rho_{+1}$ and $\rho_{-1}$. Let $\beta(X,\hat{Y})\ell(f(X),\hat{Y})$ be upper bounded by $b$. Then, for any $\delta>0$, with probability at least $1-\delta$, we have
\begin{eqnarray*}
&&\sup_{f\in F}|R_{\ell,D}(f)-\hat{R}_{\beta\ell,D_\rho}(f)|\\
&&=\sup_{f\in F}\left|E_{(X,\hat{Y})\sim D_\rho}\left[\hat{R}_{\beta\ell,D_\rho}(f)\right]-\hat{R}_{\beta\ell,D_\rho}(f)\right|\\
&&\leq \frac{1-U}{1-\rho_{-1}-\rho_{+1}}\mathfrak{R}(\ell\circ F)+b\sqrt{\frac{\log(1/\delta)}{2n}},
\end{eqnarray*}
where $U=\min_{(X,\hat{Y})}\frac{\rho_{-\hat{Y}}}{P_{D_\rho}(\hat{Y}|X)}$, and the Rademacher complexity $\mathfrak{R}(\ell\circ F)$ \cite{bartlett2003rademacher} is defined by
\begin{eqnarray*}
&&\mathfrak{R}(\ell\circ F)=E_{(X,\hat{Y})\sim D_\rho,\sigma}\left[\sup_{f\in F}\frac{2}{n}\sum_{i=1}^{n}\sigma_i\ell(f(X_i),\hat{Y}_i)\right]\ \ \
\end{eqnarray*}
and $\sigma_1,\ldots,\sigma_n$ are i.i.d. Rademacher variables.
\end{prop}

The Rademacher complexity has a convergence rate of order $\mathcal{O}(\sqrt{1/n})$ \cite{bartlett2003rademacher}. If the function class has proper conditions on its variance, the Rademacher complexity will quickly converge and is of order $\mathcal{O}(1/n)$; see, for example, \cite{bartlett2005local}. The generalization bound in Proposition \ref{rademacher} is derived using the Rademacher complexity method. Many other hypothesis complexities and methods can also be employed to derive the generalization bound.

Since
\begin{eqnarray*}
&&R_{\ell,D}(f_n)-R_{\ell,D}(f^*)\\
&&=R_{\beta\ell,D_\rho}(f_n)-R_{\beta\ell,D_\rho}(f^*)\\
&&\leq 2\sup_{f\in F}|R_{\beta\ell,D_\rho}(f)-\hat{R}_{\beta\ell,D_\rho}(f)|,\ \ \
\end{eqnarray*}
the consistency rate will therefore be inherited for learning with label noise, provided that the conditional probability $P_{D_\rho}(\hat{Y}|X)$ and noise rates $\rho_{\pm 1}$ are accurately estimated.

Based on Proposition \ref{rademacher}, we can now state our first main result for classification in the presence of label noise using our framework of importance reweighting.
\begin{thm}\label{thm1}
Any surrogate loss functions designed for the traditional classification problem can be used for classification in the presence of asymmetric RCN by employing the importance reweighting method. The consistency rate for classification with asymmetric RCN will be the same as that of the corresponding traditional classification algorithm, provided that the conditional probability $P_{D_\rho}(\hat{Y}|X)$ and noise rates $\rho_{\pm 1}$ are accurately estimated.
\end{thm}

The trade-off for using and benefitting from the abundant surrogate loss functions designed for traditional classification problems is the need to estimate the distribution $P_{D_\rho}(\hat{Y}|X)$ and noise rates $\rho_{\pm 1}$. Next, we address how to estimate the distribution and the noise rates separately.

\section{Estimating $P_{D_\rho}(\hat{Y}|X)$}\label{sec5}
We have shown that the uncertainty introduced by classification label noise can be reduced by the knowledge of weight
\begin{eqnarray*}
&&\beta(X,\hat{Y})=\frac{P_{D}(X,Y)}{P_{D_\rho}(X,\hat{Y})}=\frac{P_{D}(Y|X)}{P_{D_\rho}(\hat{Y}|X)}.
\end{eqnarray*}
In the asymmetric RCN problem,
\begin{eqnarray*}
&&\beta(X,\hat{Y})=\frac{P_{D_\rho}(\hat{Y}|X)-\rho_{-\hat{Y}}}{(1-\rho_{+1}-\rho_{-1})P_{D_\rho}(\hat{Y}|X)},
\end{eqnarray*}
and therefore the weight can be learned by using the noisy sample and the noise rates. In this section, we present three methods to estimate the conditional probability $P_{D_\rho}(\hat{Y}|X)$ with consistency analyses; how to estimate the noise rates is discussed in the next section.

\subsection{The Probabilistic Classification Method}
The conditional probability $P_{D_\rho}(\hat{Y}|X)$ can be estimated by a simple probabilistic classification method, where the corresponding link function maps the outputs of the learned predictor to the interval $[0,1]$ and thus can be interpreted as probabilities. However, such a method is parametric, which has a strong assumption that the target conditional distribution is of the form of the link function used. For example, if the logistic loss function is employed, the learned distribution will be the form of 
\[P_{D_\rho}(\hat{Y}|X,f)=\frac{1}{1+\exp{(-\hat{Y}f(X))}}.\]
When the logistic regression is correctly specified, i.e., there exists $f^*\in F$ such that $P_{D_\rho}(\hat{Y}|X,f^* )$ is equal to the target conditional distribution $P_{D_\rho}^* (\hat{Y}|X)$, the logistic regression is optimal in the sense that the approximation error is minimized (being zero). However, when the model is misspecified, which would be the case in practice, a large approximation error may be introduced even if the hypothesis class $F$ is chosen to be relatively large, which will hinder the statistical consistency for learning the target weight function $\beta^*(X,\hat{Y})$.

\begin{rema}
We found that employing the probabilistic classification method to estimate the conditional probability $P_{D_\rho}(\hat{Y}|X)$ did not perform well. Its empirical validation is therefore omitted in this paper. 
\end{rema}

\subsection{The Kernel Density Estimation Method}
In this subsection, we introduce the kernel density estimation method to estimate the conditional probability $P_{D_\rho}(\hat{Y}|X)$, which has the consistency property for learning the target weight function $\beta^*(X,\hat{Y})$.

Using Bayes' rule, we have
\begin{eqnarray}\label{bayes}
&&P_{D_\rho}(\hat{Y}|X)=\frac{P_{D_\rho}(X|\hat{Y})P_{D_\rho}(\hat{Y})}{P_{D_\rho}(X)}.
\end{eqnarray}
When the dimensionality of $\mathcal{X}$ is low and the sample size is sufficiently large, the probabilities $P_{D_\rho}(x|y),P_{D_\rho}(y)$ and $P_{D_\rho}(x)$ can be easily and efficiently estimated using the noisy sample.

If we use 
\begin{eqnarray}\label{kerneldensity0}
&&\hat{P}_{D_\rho}(\hat{Y}) = \frac{1}{n}\sum_{i=1}^{n}1_{\hat{Y}_i=\hat{Y}}
\end{eqnarray}
and the kernel density estimation method
\begin{eqnarray}\label{kerneldensity1}
&&\hat{P}_{D_\rho}(X)=\frac{1}{n}\sum_{i=1}^{n}K(X,X_i)
\end{eqnarray}
to estimate $P_{D_\rho}(\hat{Y})$ and $P_{D_\rho}(X)$, respectively (where $K(X,X_i)=k(X)k(X_i)$ is a universal kernel, see \cite{steinwart2002support}), the consistency of classification with label noise (or learning the target weight function $\beta^*(X,\hat{Y})$) is guaranteed by the following theorem.
\begin{thm}\label{consistency}
Let $\hat{P}_{D_\rho}(\hat{Y}|X)$ be an estimator for $P_{D_\rho}(\hat{Y}|X)$ using equations $(\ref{bayes})$, $(\ref{kerneldensity0})$ and $(\ref{kerneldensity1})$, and
\begin{eqnarray*}
&&\hat{\beta}(X,\hat{Y})=\frac{\hat{P}_{D_\rho}(\hat{Y}|X)-\rho_{-\hat{Y}}}{(1-\rho_{-1}-\rho_{+1})\hat{P}_{D_\rho}(\hat{Y}|X)}.
\end{eqnarray*}
Let
\begin{eqnarray*}
&&\hat{f}_{n,\hat{\beta}}=\min_{f\in F}\frac{1}{n}\sum_{i=1}^{n}\hat{\beta}(X_i,\hat{Y}_i)\ell(f(k(X_i)),\hat{Y}_i)
\end{eqnarray*}
and
\begin{eqnarray*}
&&f^*=\min_{f\in F}R[D,f,\ell(f(k(X)),{Y})].
\end{eqnarray*}
For any $\epsilon>0$, we have
\begin{eqnarray*}
&&\lim_{n\rightarrow\infty}P(R[D,\hat{f}_{n,\hat{\beta}},\ell(\hat{f}_{n,\hat{\beta}}(k(X)),{Y})]\\
&&\ \ \ \ \ \ \ \ \ \ \ \ \ \ \ \ \ \ -R[D,f^*,\ell(f^*(k(X)),{Y})]> \epsilon)=0.
\end{eqnarray*}
\end{thm}

When $P_{D_\rho}(X|\hat{Y})$ and $P_{D_\rho}(X)$ are estimated separately, although the consistency property is guaranteed by mapping features into a universal kernel induced reproducing kernel Hilbert space (RKHS), the convergence rate may be slow. Note that the kernel density estimation method is non-parametric and thus it often requires a large sample size.
Since density estimation is known to be a hard problem for high-dimensional variables, in practice, it is preferable to directly estimate the density ratio \cite{yamada2011relative} and avoid estimating the densities separately.

\subsection{The Density Ratio Estimation Method}
Density ratio estimation \cite{vapnik2013constructive} provides a way to significantly reduce the curse of dimensionality for kernel density estimation and can be estimated accurately for high-dimensional variables. Therefore, in this subsection, we introduce density ratio estimation to estimate the conditional probability distribution $P_{D_\rho}(\hat{Y}|X)$ for classification in the presence of RCN.

Three methods are frequently used for density ratio estimation, including the moment matching approach, the probabilistic classification approach and the ratio matching approach; see \cite{sugiyama2010density}. Since the probabilistic classification approach may introduce a large approximation error, in practice, the moment matching and ratio matching methods are more preferable \cite{kanamori2010theoretical}, where the density ratio $r(X)=P_1 (X)/P_2 (X)$ can be modelled by employing linear or non-linear functions. If proper reproducing kernel Hilbert spaces are chosen to be the hypothesis classes, the approximation errors of the moment matching and ratio matching methods could be small. Although these methods introduce approximation errors for learning the weight $\beta^* (X,\hat{Y})$, their efficiency has been widely and empirically proven \cite{huang2006correcting,kanamori2009least,sun2011two}.

In this paper, we exploit the ratio matching approach that employs the Bregman divergence \cite{nock2009bregman} (KLIEP \cite{sugiyama2008direct}) to estimate the conditional probability distribution $P_{D_\rho}(\hat{Y}|X)$. 
It is proven that the ratio matching approach exploiting the Bregman divergence \cite{nock2009bregman} is consistent with the optimal approximation in the hypothesis class\footnote{Parametric modeling is used for estimating density ratio. We provide the proof of consistency in Section \ref{proofr}.}. 

The following theorem provides an assurance that our importance reweighting method that exploits density ratio estimation is consistent.
\begin{thm}\label{densityratio}
When employing the density ratio estimation method to estimate the conditional probability distribution $P_{D_\rho}(\hat{Y}|X)$ $($and $\beta(X,\hat{Y}))$, if the hypothesis class for estimating the density ratio is chosen properly so that the approximation error is zero, for any $\epsilon>0$, we have
\begin{eqnarray*}
&&\lim_{n\rightarrow\infty}P(R[D,\hat{f}_{n,\hat{\beta}},\ell(\hat{f}_{n,\hat{\beta}}(X),Y)]\\
&&\ \ \ \ \ \ \ \ \ \ \ \ \ \ \ \ \ \ -R[D,f^*,\ell(f^*(X),Y)]> \epsilon)=0,
\end{eqnarray*}
where $\hat{\beta}(X,\hat{Y})$ is the same as that defined in Theorem \ref{consistency}, $\hat{f}_{n,\hat{\beta}}=\min_{f\in F}\frac{1}{n}\sum_{i=1}^{n}\hat{\beta}(X_i,\hat{Y}_i)\ell(f(X_i),\hat{Y}_i)$ and $f^*=\min_{f\in F}R[D,f,\ell(f(X),{Y})]$.
\end{thm}

The convergence rate is characterized in the following proposition.
\begin{prop}\label{densityratiorate}
Under the settings of Theorem \ref{densityratio}, if the Bregman divergence degenerates to square distance, for any $\delta>0$, with probability at least $1-3\delta$, the following holds:
\begin{eqnarray*}
&R[D,\hat{f}_{n,\hat{\beta}},\ell(\hat{f}_{n,\hat{\beta}}(X),Y)]-R[D,f^*,\ell(f^*(X),Y)]\\
&\leq \mathcal{O}\left(\mathfrak{R}(\ell\circ F)+\sqrt{\frac{\log(1/\delta)}{n}}+ \sqrt{\mathfrak{R}_{\text{BDR}}+\sqrt{\frac{\log(1/\delta)}{n}}}\right),
\end{eqnarray*}
where $\mathfrak{R}_{\text{BDR}}$ is the Rademacher complexity induced by estimating the density ratio using square distance and is defined in Section \ref{proofr}.
\end{prop}

The proofs of Theorem \ref{densityratio} and Proposition \ref{densityratiorate} are provided in the supplementary material.

The convergence rate in Proposition \ref{densityratiorate} could be a certain rate of order $\mathcal{O}(1/\min(n_+,n_-)^{1/4})$ because $\mathfrak{R}_{\text{SDR}}\leq\mathcal{O}(\sqrt{1/\min(n_+,n_-)})$, where $n_+$ and $n_-$ denote the number of positive labels and negative labels of the noisy sample, respectively.


\section{Estimating the Noise Rates}\label{sec6}
Most existing algorithms designed for RCN problems need the knowledge of the noise rates. Scott et al. \cite{conf/colt/ScottBH13,blanchard2010semi} developed lower bounds for the inversed noise rates $\pi_{+1} = P(Y=-1|\hat{Y}=+1)$ and $\pi_{-1} = P(Y=+1|\hat{Y}=-1)$, under the irreducibility assumption, which are consistent with the target inversed noise rates and can therefore be used as estimators for the inversed noise rates. However, the convergence rate could be slow. Then, during the preparation of this manuscript, Scott \cite{scott2015a} released an efficient implementation to estimate the inversed noise rates and introduced the distributional assumption $\{(X,Y)|Y=1\}\not\subset\{(X,Y)|Y=-1\}$ and $\{(X,Y)|Y=-1\}\not\subset\{(X,Y)|Y=1\}$ to the label noise classification problem. The distributional assumption is sufficient for the irreducibility assumption and thus is slightly stronger. Scott then proved that the distributional assumption ensures an asymptotic convergence rate of order $\mathcal{O}(\sqrt{1/n})$ for estimating the inversed noise rates.  

To the best of our knowledge, no efficient method has been proposed to estimate the noise rates and how to estimate them remains an open problem \cite{yang2012multiple}. We first provide upper bounds for the noise rates and show that with a mild assumption on the ``clean" data, they can be used to efficiently estimate the noise rates.
\begin{thm}\label{thm4}
We have that
\begin{eqnarray*}
&&\rho_{\hat{Y}}\leq P_{D_\rho}(-\hat{Y}|X).
\end{eqnarray*}
Moreover, if the assumption holds that there exists $x_{-1},x_{+1}\in\mathcal{X}$, such that $P_D({Y}=+1|x_{-1})=P_D({Y}=-1|x_{+1})=0$, we have
\begin{eqnarray*}
&&\rho_{-1}=P_{D_\rho}(\hat{Y}=+1|x_{-1})
\end{eqnarray*}
and
\begin{eqnarray*}
&&\rho_{+1}=P_{D_\rho}(\hat{Y}=-1|x_{+1}),
\end{eqnarray*}
which means
\begin{eqnarray*}
&&{\rho}_{-\hat{Y}}=\min_{X\in\mathcal{X}}P_{D_\rho}(\hat{Y}|X).
\end{eqnarray*}
\end{thm}

Theorem \ref{thm4} shows that under the assumption that there exists $x_{-1},x_{+1}\in\mathcal{X}$, such that $P_D({Y}=+1|x_{-1})=P_D({Y}=-1|x_{+1})=0$, $\min_{X\in\mathcal{X}}P_{D_\rho}(\hat{Y}|X)$ is a consistent estimator for the noise rates. The convergence rate for estimating the noise rates is the same as that of estimating the conditional distribution $P_{D_\rho} (\hat{Y}|X)$. We therefore could obtain fast convergence rates for estimating the noise rates via finite sample analysis. For example, if the hypothesis class has proper conditions on its variance, the Rademacher complexity will quickly converge and is of order $\mathcal{O}(1/n)$ \cite{bartlett2005local}.
\begin{rema}
We have proven the consistency property of the joint estimation of the weight and classifier in Theorems \ref{consistency} and \ref{densityratio}, and characterized the convergence rates of the joint estimation in Proposition \ref{densityratiorate}. According to Theorem \ref{thm4}, the results can be easily extended to the joint estimation of the weight, noise rate and classifier of our importance reweighting method. We provide detailed proofs in the supplementary material.
\end{rema}

For classification problems, the assumption in Theorem \ref{thm4} can be easily held. If an observation $x\in\mathcal{X}$ is far from the target classifier, it is likely that the conditional probability $P_D(y=+1|x)$ (or $P_D(y=-1|x)$) is equal to zero or very small. With the assumption that there exist $x_{-1},x_{+1}\in\mathcal{X}$ such that $P_D(y=+1|x_{-1})$ and $P_D(y=-1|x_{+1})$ are very small, we can efficiently estimate $\rho_{y}$ by
\begin{eqnarray*}
&&\min_{X\in\mathcal{X}}P_{D_\rho}(\hat{Y}|X).
\end{eqnarray*}
In our experiments, we estimate $\rho_{y}$ by
\begin{eqnarray*}
&&\hat{\rho}_{-\hat{Y}}=\min_{X\in\{X_1,\ldots,X_n\}}\hat{P}_{D_\rho}(\hat{Y}|X).
\end{eqnarray*}

It is not hard to see that Scott's distributional assumption is equal to ours in Theorem \ref{thm4}. Interestingly, this is not a coincidence. In Proposition 3 of \cite{scott2015a}, Scott derived that
\begin{align*}
&P_{D_{\rho}}(X|\hat{Y})\\ 
&=\left(1-\frac{\pi_{\hat{Y}}}{1-\pi_{-\hat{Y}}}\right)P_D(X|Y)+\frac{\pi_{\hat{Y}}}{1-\pi_{-\hat{Y}}}P_{D_{\rho}}(X|-\hat{Y}).
\end{align*}
Note that $\left(1-\frac{\pi_{\hat{Y}}}{1-\pi_{-\hat{Y}}}\right)P_D(X|Y)\geq 0$. Thus, $\frac{\pi_{\hat{Y}}}{1-\pi_{-\hat{Y}}}$ can be consistently estimated by $\min_{X\in\mathcal{X}} \frac{P_{D_{\rho}}(X|\hat{Y})}{P_{D_{\rho}}(X|-\hat{Y})}$ if there exists an $x\in \mathcal{X}$ such that $P_D(x|Y)=0$, where $\frac{P_{D_{\rho}}(X|\hat{Y})}{P_{D_{\rho}}(X|-\hat{Y})}$ is the slope to the point $(1,1)$ in the receiver operating characteristic (ROC) space defined in \cite{conf/colt/ScottBH13,scott2015a}. In the proof of Theorem \ref{thm4}, we also derived that
\begin{eqnarray*}
&&P_{D_{\rho}}(\hat{Y}|X)=(1-\rho_{-1}-\rho_{+1})P_D(Y|X)+\rho_{-\hat{Y}}.
\end{eqnarray*}
Since $(1-\rho_{-1}-\rho_{+1})P_D(Y|X)$ is non-negative, our estimator ${\rho}_{-\hat{Y}}=\min_{X\in\mathcal{X}}P_{D_\rho}(\hat{Y}|X)$ is consistent based on the assumption that there exists an $x\in X$ such that $P_D (Y|x)=0$. Having the above knowledge in mind, we can improve the theoretical analysis for estimation the inversed noise rates in \cite{conf/colt/ScottBH13,scott2015a} (and the mixture proportion estimation) by employing finite sample analysis.

We can design importance reweighting algorithms for classification with label noise by employing the inversed noise rates.
\begin{lema}\label{weight1}
When using the importance reweighting method to address the asymmetric RCN problem, the weight given to a noisy example $(X,\hat{Y})\sim D_\rho$ can be derived by exploiting the inversed noise rates:
\begin{eqnarray*}
&\beta(X,\hat{Y})&=\frac{P_{D}(Y|X)}{P_{D_\rho}(\hat{Y}|X)}\\
&&=\frac{(1-\pi_{-1}-\pi_{+1})P_{D_\rho}(\hat{Y}|X)+\pi_{-\hat{Y}}}{P_{D_\rho}(\hat{Y}|X)}.
\end{eqnarray*}
The weight $\beta(X,\hat{Y})$ is non-negative\footnote{The inversed noise rates are defined so that $\pi_{-1}+\pi_{+1}\leq 1$, see, \cite{scott2015a}.} if $P_{D_\rho}(\hat{Y}|X)\neq 0$. If $P_{D_\rho}(\hat{Y}|X)= 0$, we intuitively let $\beta(X,\hat{Y})= 0$.
\end{lema}

\begin{rema}
We employed Scott's method \cite{scott2015a} to estimate the inversed noise rates and found that the importance reweighting method exploiting the estimated inversed noise rates did not perform well, so the results are omitted. There might be two reasons which could possibly explain the poor performance: (1) Scott's estimator $\frac{\pi_{\hat{Y}}}{1-\pi_{-\hat{Y}}}=\min_{X\in\mathcal{X}} \frac{P_{D_{\rho}}(X|\hat{Y})}{P_{D_{\rho}}(X|-\hat{Y})}$ has the form of density ratio estimation, and is more complex than our estimator ${\rho}_{-\hat{Y}}=\min_{X\in\mathcal{X}}P_{D_\rho}(\hat{Y}|X)$, which has the form of the conditional distribution. (2) How to choose the kernel width to obtain the ROC in Scott's method has remained elusive.
\end{rema}

\section{Proof}\label{proofsection}
In this section, we provide detailed proofs of the assertions made in previous sections.
\subsection{Proof of Lemma \ref{weight}}\label{proof}

For label noise problem, we have shown that
\begin{eqnarray*}
\beta(X,\hat{Y})=\frac{P_{D}(Y|X)}{P_{D_\rho}(\hat{Y}|X)}.
\end{eqnarray*}
When the label noise is of asymmetric RCN, we have
\begin{eqnarray*}
&&P_{D_\rho}(+1|X)\\
&&=P(\hat{Y}=+1,Y=+1|X)+P(\hat{Y}=+1,Y=-1|X)\\
&&=P(\hat{Y}=+1|Y=+1,X)P_D(Y=+1|X)\\
&&\ \ \ +P(\hat{Y}=+1|Y=-1,X)P_D(Y=-1|X)\\
&&=P(\hat{Y}=+1|Y=+1)P_D(Y=+1|X)\\
&&\ \ \ +P(\hat{Y}=+1|Y=-1)P_D(Y=-1|X)\\
&&=(1-\rho_{+1})P_D(Y=+1|X)\\ 
&&\ \ \ +\rho_{-1}(1-P_D(Y=+1|X))\\
&&=(1-\rho_{-1}-\rho_{+1})P_D(Y=+1|X)+\rho_{-1}\\
&&\geq \rho_{-1}. 
\end{eqnarray*}
Similarly, it gives
\begin{eqnarray*}
&P_{D_\rho}(-1|X)&=(1-\rho_{-1}-\rho_{+1})P_D(Y=-1|X)+\rho_{+1}\\ 
&&\geq \rho_{+1}.
\end{eqnarray*}
We therefore have
\begin{eqnarray*}
&&P_D(Y|X)=\frac{P_{D_\rho}(\hat{Y}|X)-\rho_{-\hat{Y}}}{(1-\rho_{-1}-\rho_{+1})}.
\end{eqnarray*}
Thus,
\begin{eqnarray*}
&\beta(X,\hat{Y})&=\frac{P_{D}(Y|X)}{P_{D_\rho}(\hat{Y}|X)}\\
&&=\frac{P_{D_\rho}(\hat{Y}|X)-\rho_{-\hat{Y}}}{(1-\rho_{+1}-\rho_{-1})P_{D_\rho}(\hat{Y}|X)}.
\end{eqnarray*}
We intuitively let $\beta(X,\hat{Y})= 0$, if $P_{D_\rho}(\hat{Y}|X)=0$. Since $P_{D_\rho}(\hat{Y}|X)\geq\rho_{-\hat{Y}}$, we can conclude that $\beta(X,\hat{Y})\geq 0$.

\hfill$\blacksquare$

\subsection{Proofs of Proposition \ref{rademacher} and Theorem \ref{thm1}}
We start by introducing the Rademacher complexity method \cite{bartlett2003rademacher} for deriving generalization bounds.

Let $\sigma_1,\ldots,\sigma_n$ be independent Rademacher variables, $X_1,\ldots,X_n$ be i.i.d. variables and $F$ be a real-valued function class. The Rademacher complexity of the function class over the variable is defined as 
\[\mathfrak{R}(F)=E_{X,\sigma}\left[\sup_{f\in F}\frac{2}{n}\sum_{i=1}^{n}\sigma_if(X_i)\right].\]

\begin{thm}[\cite{bartlett2003rademacher}]\label{upperR}
Let $F$ be a real-valued function class on $\mathcal{X}$, $S=\{X_1,\ldots,X_n\}\in\mathcal{X}^n$
and 
\[\Phi(S)=\sup_{f\in F}\left|\frac{1}{n}\sum_{i=1}^{n}E[f(X)]-f(X_i)\right|.\]
Then, $E_{S}[\Phi(S)]\leq \mathfrak{R}(F)$.
\end{thm}

The following theorem, proven utilizing Theorem \ref{upperR} and Hoeffding's inequality, plays an important role in deriving the generalization bounds.
\begin{thm}[\cite{bartlett2003rademacher}]\label{generR}
Let $F$ be an $[a,b]$-valued function class on $\mathcal{X}$, and $S=\{X_1,\ldots,X_n\}\in\mathcal{X}^n$. Then, for any $f\in F$ and any $\delta>0$, with probability at least $1-\delta$, we have
\begin{eqnarray*}
E_X[f(X)]-\frac{1}{n}\sum_{i=1}^{n}f(X_i)\leq \mathfrak{R}(F)+(b-a)\sqrt{\frac{\log(1/\delta)}{2n}}.
\end{eqnarray*}
\end{thm}

According to Theorem \ref{generR}, we can easily prove that for any $[0,b]$-valued function class and $\delta>0$, with probability at least $1-\delta$, the following holds
\begin{eqnarray*}
&&\sup_{f\in F}|E_{(X,\hat{Y})\sim D_\rho}\hat{R}_{\beta\ell,D_\rho}-\hat{R}_{\beta\ell,D_\rho}|\\ 
&&\leq \mathfrak{R}(\beta\circ\ell\circ F)+b\sqrt{\frac{\log(1/\delta)}{2n}}.
\end{eqnarray*}
Since $\beta$ is upper bounded by
\begin{eqnarray*}
&&\frac{1-U}{1-\rho_{-1}-\rho_{+1}},
\end{eqnarray*}
where
\begin{eqnarray*}
&&U=\min_{(X,\hat{Y})}\frac{\rho_{-\hat{Y}}}{P_{D_\rho}(\hat{Y}|X)},
\end{eqnarray*}
using the Lipschitz composition property of Rademacher complexity, which is also known as the Talagrand's Lemma (see, e.g., Lemma 4.2 in \cite{mohri2012foundations}), we have
\begin{eqnarray*}
\mathfrak{R}(\beta\circ\ell\circ F)\leq\frac{1-U}{1-\rho_{-1}-\rho_{+1}}\mathfrak{R}(\ell\circ F).
\end{eqnarray*}
Propostion \ref{rademacher} can be proven together with the fact that $E_{(X,\hat{Y})\sim D_\rho}[\hat{R}_{\beta\ell,D_\rho}]=R_{\beta\ell,D_\rho}=R_{\ell,D}$.\hfill$\blacksquare$

Theorem \ref{thm1} follows from Proposition \ref{rademacher}.
\subsection{Proof of Theorem \ref{consistency}}
We begin with the following lemma.
\begin{lema}\label{le}
Let $K(X_1,X_2)=k(X_1)k(X_2)$ be a universal kernel, where $k:\mathcal{X}\rightarrow\mathcal{H}$ is a feature map into a feature space. Let
\[\hat{P}_{D_\rho}(\hat{Y}) = \frac{1}{n}\sum_{i=1}^{n}1_{\hat{Y}_i=\hat{Y}}\]
and
\[\hat{P}_{D_\rho}(X)=\frac{1}{n}\sum_{i=1}^{n}K(X,X_i).\]
Then, $\hat{P}_{D_\rho}(\hat{Y})$ and $\hat{P}_{D_\rho}(X)$ will converge to their target distributions $P_{D_\rho}(\hat{Y})$ and $P_{D_\rho}(X)$ in the induced RKHS $\mathcal{H}$, respectively.
\end{lema}
The proof relies on the following theorem proven by Gretton et al. \cite{gretton2009covariate}.
\begin{thm}\label{conv}
Let $\mathcal{P}$ be the space of all probability distributions on an RKHS $\mathcal{H}$ induced by a universal kernel $K(X_1,X_2)=k(X_1)k(X_2)$. Define $\mu:\mathcal{P}\rightarrow\mathcal{H}$ as the expectation operator that $\mu(P)=E_{X\sim P(X)}[k(X)]$. The operator $\mu$ is a bijection between $\mathcal{P}$ and $\{\mu(P)|P\in\mathcal{P}\}$.
\end{thm}

\textbf{Proof of Lemma \ref{le}.}
Since
\begin{eqnarray*}
&&E[\hat{P}_{D_\rho}(\hat{Y})]= \frac{1}{n}\sum_{i=1}^{n}E1_{\hat{Y}_i=\hat{Y}}=P_{D_\rho}(\hat{Y}),
\end{eqnarray*}
using the weak law of large numbers, for any $\epsilon>0$, we have
\begin{eqnarray*}
&&\lim_{n\rightarrow\infty}P\left(|\hat{P}_{D_\rho}(\hat{Y})-P_{D_\rho}(\hat{Y})|\geq\epsilon\right)=0.
\end{eqnarray*}
So, $\hat{P}_{D_\rho}(\hat{Y})$ will converge to its target distribution $P_{D_\rho}(\hat{Y})$.

We then prove that $\hat{P}_{D_\rho}(X)$ converges to $P_{D_\rho}(X)$ in the RKHS by using Theorem \ref{conv} and showing that
\begin{eqnarray*}
\int\hat{P}_{D_\rho}(X)k(X)dX=E_{X\sim P_{D_\rho}(X)}[k(X)], \ \text{when}\ n\rightarrow\infty.
\end{eqnarray*}

We have that
\begin{eqnarray*}
&\hat{P}_{D_\rho}(X)&=\frac{1}{n}\sum_{i=1}^{n}K(X,X_i)=\frac{1}{n}\sum_{i=1}^{n}k(X)k(X_i)\\
&&=\frac{k(X)}{n}\sum_{i=1}^{n}k(X_i).
\end{eqnarray*}
By properly modifying the kernel map $k$ by a constant so that $\int k^2(X)dX=1$, we have
\begin{eqnarray*}
&\int\hat{P}_{D_\rho}(X)k(X)dX&=\frac{1}{n}\sum_{i=1}^{n}k(X_i)\int k^2(X)dX\\
&&=\frac{1}{n}\sum_{i=1}^{n}k(X_i).
\end{eqnarray*}
Moreover, for any $\epsilon>0$, using Hoeffding's inequality, the following holds
\begin{eqnarray*}
\lim_{n\rightarrow\infty}P\left(\left|\frac{1}{n}\sum_{i=1}^{n}k(X_i)-E_{X\sim P_{D_\rho}(X)}[k(X)]\right|\geq\epsilon\right)=0.
\end{eqnarray*}
By combing the above two equations, we can conclude that
\begin{eqnarray*}
&&\lim_{n\rightarrow\infty}P\left(\left|\int\hat{P}_{D_\rho}(X)k(X)dX\right.\right.\\
&&\left.\left.\ \ \ \ \ \ \ \ \ \ \ \ \ \ \ \ \ \ \ \ \ \ \  \ -E_{X\sim P_{D_\rho}(X)}[k(X)]\right|\geq\epsilon\right)=0.
\end{eqnarray*}
According to Theorem \ref{conv}, we have that the estimator $\hat{P}_{D_\rho}(x)$ will converge to $P_{D_\rho}(X)$ in $\mathcal{H}$.\hfill$\blacksquare$

\textbf{Proof of Theorem \ref{consistency}.} In the universal kernel induced RKHS, we have proven that
\begin{eqnarray*}
&&\hat{P}_{D_\rho}(\hat{Y})=P_{D_\rho}(\hat{Y}),\ \text{when}\ n\rightarrow\infty
\end{eqnarray*}
and
\begin{eqnarray*}
&&\hat{P}_{D_\rho}(X)=P_{D_\rho}(X),\ \text{when}\ n\rightarrow\infty.
\end{eqnarray*}
Thus, we have
\begin{eqnarray}\label{beta}
&&\hat{\beta}(X,\hat{Y})=\beta(X,\hat{Y}),\ \text{when}\ n\rightarrow\infty.
\end{eqnarray}
In Proposition \ref{rademacher}, we have proven that
\begin{eqnarray}\label{gener}
&&\sup_{f\in F}|R[D_\rho,f,\beta(X,\hat{Y})\ell(f(X),\hat{Y})]\nonumber\\
&& \ \ \ \ \ \ \ \ \ \ \ \ -\hat{R}[D_\rho,f,\beta(X,\hat{Y})\ell(f(X),\hat{Y})]|=0,\nonumber\\
&& \ \ \ \ \ \ \ \ \ \ \ \ \ \ \ \ \ \ \ \ \   \ \ \  \ \ \ \ \ \ \ \ \ \ \text{when}\ n\rightarrow\infty.\ \ \ \ \ \
\end{eqnarray}
By substitution from equation (\ref{beta}) into equation (\ref{gener}), we have
\begin{eqnarray}\label{gener2}
&&\sup_{f\in F}|R[D_\rho,f,\hat{\beta}(X,\hat{Y})\ell(f(X),\hat{Y})]\nonumber\\
&&\ \ \ \ \ \ \ \ \ \ \ \ -\hat{R}[D_\rho,f,\hat{\beta}(X,\hat{Y})\ell(f(X),\hat{Y})]|=0,\nonumber\\
&& \ \ \ \ \ \ \ \ \ \ \ \ \ \ \ \ \ \ \ \ \  \ \ \ \ \  \ \ \ \ \ \ \ \ \ \ \text{when}\ n\rightarrow\infty.\ \ \ \ \ \
\end{eqnarray}
Let
\begin{eqnarray*}
&&\hat{f}_{n,\hat{\beta}}=\min_{f\in F}\frac{1}{n}\sum_{i=1}^{n}\hat{\beta}(X_i,\hat{Y}_i)\ell(f(k(X_i)),\hat{Y}_i)
\end{eqnarray*}
and
\begin{eqnarray*}
&&f^*=\min_{f\in F}R[D,f,\ell(f(k(X)),Y)].
\end{eqnarray*}

\begin{figure*}[t]
\hrule
\begin{eqnarray}\label{gener3}
&&R[D_{\rho},\hat{f}_{n,\hat{\beta}},\hat{\beta}(X,\hat{Y})\ell(\hat{f}_{n,\hat{\beta}}(k(X)),\hat{Y})]-R[D_{\rho},f^*,\hat{\beta}(X,\hat{Y})\ell(f^*(k(X)),\hat{Y})]\nonumber\\
&&=R[D_\rho,\hat{f}_{n,\hat{\beta}},\hat{\beta}(X,\hat{Y})\ell(\hat{f}_{n,\hat{\beta}}(k(X)),\hat{Y})]-\hat{R}[D_\rho,\hat{f}_{n,\hat{\beta}},\hat{\beta}(X,\hat{Y})\ell(\hat{f}_{n,\hat{\beta}}(k(X)),\hat{Y})]\nonumber\\
&&\ \ \ +\hat{R}[D_\rho,\hat{f}_{n,\hat{\beta}},\hat{\beta}(X,\hat{Y})\ell(\hat{f}_{n,\hat{\beta}}(k(X)),\hat{Y})]-\hat{R}[D_\rho,f^*,\hat{\beta}(X,\hat{Y})\ell(f^*(k(X)),\hat{Y})]\nonumber \\
&&\ \ \ +\hat{R}[D_\rho,f^*,\hat{\beta}(X,\hat{Y})\ell(f^*(k(X)),\hat{Y})]-R[D_\rho,f^*,\hat{\beta}(X,\hat{Y})\ell(f^*(k(X)),\hat{Y})]\nonumber \\
&&\leq R[D_\rho,\hat{f}_{n,\hat{\beta}},\hat{\beta}(X,\hat{Y})\ell(\hat{f}_{n,\hat{\beta}}(k(X)),\hat{Y})]-\hat{R}[D_\rho,\hat{f}_{n,\hat{\beta}},\hat{\beta}(X,\hat{Y})\ell(\hat{f}_{n,\hat{\beta}}(k(X)),\hat{Y})]\nonumber\\
&&\ \ \ +\hat{R}[D_\rho,f^*,\hat{\beta}(X,\hat{Y})\ell(f^*(k(X)),\hat{Y})]-R[D_\rho,f^*,\hat{\beta}(X,\hat{Y})\ell(f^*(k(X)),\hat{Y})]\nonumber \\
&&\leq 2\sup_{f\in F}|\hat{R}[D_\rho,f,\hat{\beta}(X,\hat{Y})\ell(f(k(X)),\hat{Y})]-R[D_\rho,f,\hat{\beta}(X,\hat{Y})\ell(f(k(X)),\hat{Y})]|.
\end{eqnarray}
\hrule
\begin{eqnarray}\label{comr}
&&\mathfrak{R}_{\text{BDR}}=E_{X\sim D_\rho,\sigma}\left[\frac{2}{n_2}\sum_{i=1}^{n_2}\sigma_i\nabla f(r(X_i^{\text{de}}))r(X_i^{\text{de}})-\frac{2}{n_2}\sum_{i=1}^{n_2} \sigma_if(r(X_i^{\text{de}}))-\frac{2}{n_1}\sum_{i=1}^{n_1}\sigma_i\nabla f(r(X_i^{\text{nu}}))\right].\ \ \
\end{eqnarray}
\hrule
\end{figure*}

We have inequalities (\ref{gener3}). The first inequality in inequalities (\ref{gener3}) holds because of the definition of $\hat{f}_{n,\hat{\beta}}$.

For sufficiently large $n$, using equations (\ref{beta}), (\ref{gener2}) and (\ref{gener3}), we have
\begin{eqnarray*}\label{gener4}
&&R[D,\hat{f}_{n,\hat{\beta}},\ell(\hat{f}_{n,\hat{\beta}}(k(X)),Y)]\\
&&\ \ \ -R[D,f^*,\ell(f^*（k(X)),Y)]\\
&&=R[D_{\rho},\hat{f}_{n,\hat{\beta}},\beta(X,\hat{Y})\ell(\hat{f}_{n,\hat{\beta}}(k(X)),\hat{Y})]\\
&&\ \ \ -R[D_{\rho},f^*,\beta(X,\hat{Y})\ell(f^*(k(X)),\hat{Y})]\\
&&=R[D_{\rho},\hat{f}_{n,\hat{\beta}},\hat{\beta}(X,\hat{Y})\ell(\hat{f}_{n,\hat{\beta}}(k(X)),\hat{Y})]\\
&&\ \ \ -R[D_{\rho},f^*,\hat{\beta}(X,\hat{Y})\ell(f^*(k(X)),\hat{Y})]\\
&&\leq 2\sup_{f\in F}\left|\hat{R}[D_\rho,f,\hat{\beta}(X,\hat{Y})\ell(f(k(X)),\hat{Y})]\right.\\
&&\ \ \ \left.-R[D_\rho,f,\hat{\beta}(X,\hat{Y})\ell(f(k(X)),\hat{Y})]\right|\\
&&=0.
\end{eqnarray*}
This concludes the proof of Theorem \ref{consistency}.\hfill$\blacksquare$
\subsection{Proof of Theorem \ref{thm4}}
In the proof of Lemma \ref{weight}, we have proven that
\begin{eqnarray*}
P_{D_\rho}(+1|X)=(1-\rho_{-1}-\rho_{+1})P_D(Y=+1|X)+\rho_{-1},
\end{eqnarray*}
If there exists $x_{-1}\in\mathcal{X}$ such that
\begin{eqnarray*}
&&P_D(Y=+1|x_{-1})=0,
\end{eqnarray*}
then
\begin{eqnarray*}
&&P_{D_\rho}(\hat{Y}=+1|x_{-1})=\rho_{-1}.
\end{eqnarray*}
Similarly, 
\[P_{D_\rho}(-1|X)=(1-\rho_{-1}-\rho_{+1})P_D(Y=-1|X)+\rho_{+1},\] 
and if there exists $x_{+1}\in\mathcal{X}$ such that
\begin{eqnarray*}
&&P_D(Y=-1|x_{+1})=0,
\end{eqnarray*}
which means
\begin{eqnarray*}
&&P_{D_\rho}(\hat{Y}=-1|x_{+1})=\rho_{+1}.
\end{eqnarray*}
We therefore have
\begin{eqnarray*}
&&{\rho}_{-\hat{Y}}=\min_{X\in\mathcal{X}}P_{D_\rho}(\hat{Y}|X),
\end{eqnarray*}
which concludes the proof. \hfill$\blacksquare$

\subsection{Proof of Lemma \ref{weight1}}
Similar to the proof of Lemma \ref{weight}, we have
\begin{eqnarray*}
&&P_{D}(+1|X)\\
&&=P({Y}=+1,\hat{Y}=+1|X)+P({Y}=+1,\hat{Y}=-1|X)\\
&&=P({Y}=+1|\hat{Y}=+1,X)P_{D_\rho}(\hat{Y}=+1|X)\\
&&\ \ \ +P({Y}=+1|\hat{Y}=-1,X)P_{D_\rho}(\hat{Y}=-1|X)\\
&&=P({Y}=+1|\hat{Y}=+1)P_{D_\rho}(\hat{Y}=+1|X)\\
&&\ \ \ +P({Y}=+1|\hat{Y}=-1)P_{D_\rho}(\hat{Y}=-1|X)\\
&&=(1-\pi_{+1})P_{D_\rho}(\hat{Y}=+1|X)\\ 
&&\ \ \ +\pi_{-1}(1-P_{D_\rho}(\hat{Y}=+1|X))\\
&&=(1-\pi_{-1}-\pi_{+1})P_{D_\rho}(\hat{Y}=+1|X)+\pi_{-1}\\
&&\geq \pi_{-1}. 
\end{eqnarray*}
We also have
\begin{eqnarray*}
&&P_{D}(-1|X)\\ 
&&=(1-\pi_{-1}-\pi_{+1})P_{D_\rho}(\hat{Y}=-1|X)+\pi_{+1}\\
&&\geq \pi_{+1}. 
\end{eqnarray*}
Thus,
\begin{eqnarray*}
&\beta(X,\hat{Y})&=\frac{P_{D}(Y|X)}{P_{D_\rho}(\hat{Y}|X)}\\
&&=\frac{(1-\pi_{-1}-\pi_{+1})P_{D_\rho}(\hat{Y}|X)+\pi_{-\hat{Y}}}{P_{D_\rho}(\hat{Y}|X)}.
\end{eqnarray*}
We intuitively let $\beta(X,\hat{Y})= 0$, if $P_{D_\rho}(\hat{Y}|X)=0$. Then, we can conclude that $\beta(X,\hat{Y})\geq 0$.\hfill$\blacksquare$

\subsection{Consistency of Density Ratio Estimation}\label{proofr}
We first introduce how to use the ratio matching method under the Bregman divergence to estimate
\begin{eqnarray*}
&&r^*(X)=\frac{P_{D_\rho}(X|\hat{Y})}{P_{D_\rho}(X)}.
\end{eqnarray*}
The discrepancy from the true density ratio $r^*$ to a density ratio model $r$ measured by the Bregman divergence (BD) is as follows:
\begin{eqnarray*}
&&\text{BD}_f(r^*\|r)=\int P_{D_\rho}(X)\left\{f(r^*(X))-f(r(X))\right.\\
&&\left.\ \ \ \ \ \ \ \ \ \ \ \ \ \ \ \ \ \ \ \ \ -\nabla f(r(X))(r^*(X)-r(X))\right\}dX,
\end{eqnarray*}
where $f$ is a convex function and $\nabla f(X)$ denotes the subgradient of $f(X)$.

Let $X_1^{\text{nu}},\ldots,X_{n_1}^{\text{nu}}$ be the i.i.d. sample of the numerator distribution and $X_1^{\text{de}},\ldots,X_{n_2}^{\text{de}}$ the i.i.d. sample of the denominator distribution. An empirical approximation of $\text{BD}_f(r^*\|r)$ is given by
\begin{eqnarray*}
&\hat{\text{BD}}_f(r^*\|r)&=\frac{1}{n_2}\sum_{i=1}^{n_2}\nabla f(r(X_i^{\text{de}}))r(X_i^{\text{de}})\\
&&-\frac{1}{n_2}\sum_{i=1}^{n_2} f(r(X_i^{\text{de}}))-\frac{1}{n_1}\sum_{i=1}^{n_1}\nabla f(r(X_i^{\text{nu}})).
\end{eqnarray*}
Let
\begin{eqnarray*}
&&\hat{r}(X)=\arg\min_r\hat{\text{BD}}_f(r^*\|r)
\end{eqnarray*}
and
\begin{eqnarray*}
&&r'=\arg\min_r\text{BD}_f(r^*\|r).
\end{eqnarray*}
If the hypothesis class includes $r^*$, we have
\begin{eqnarray*}
&&\text{BD}_f(r^*\|\hat{r})\\
&&=\text{BD}_f(r^*\|\hat{r})-\text{BD}_f(r^*\|r')\\
&&=\text{BD}_f(r^*\|\hat{r})-\hat{\text{BD}}_f(r^*\|\hat{r})+\hat{\text{BD}}_f(r^*\|\hat{r})\\
&&\ \ \ -\hat{\text{BD}}_f(r^*\|r')+\hat{\text{BD}}_f(r^*\|r')-\text{BD}_f(r^*\|r')\\
&&\leq\text{BD}_f(r^*\|\hat{r})-\hat{\text{BD}}_f(r^*\|\hat{r})\\
&&\ \ \ +\hat{\text{BD}}_f(r^*\|r')-\text{BD}_f(r^*\|r')\\
&&\leq 2\sup_r|\hat{\text{BD}}_f(r^*\|r)-\text{BD}_f(r^*\|r)|,
\end{eqnarray*}
where the first inequality holds because of the definition of $\hat{r}$.

\begin{figure*}[t]
  \begin{center}
 \scalebox{0.8}{\includegraphics[width=.55\textwidth, height=0.43\textwidth]{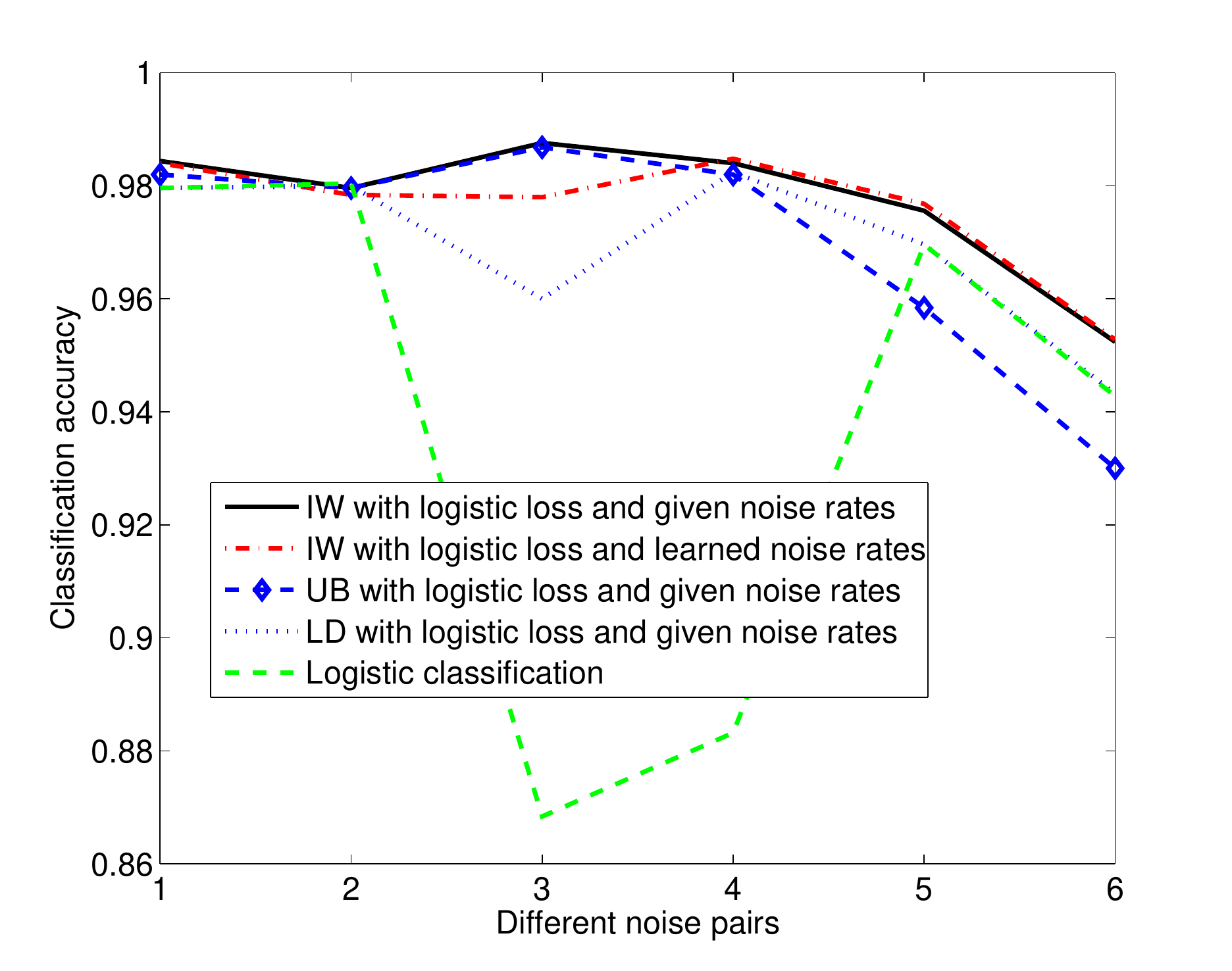}\ \ \ \ \ \ \ \ \ \  \includegraphics[width=.55\textwidth,height=0.43\textwidth]{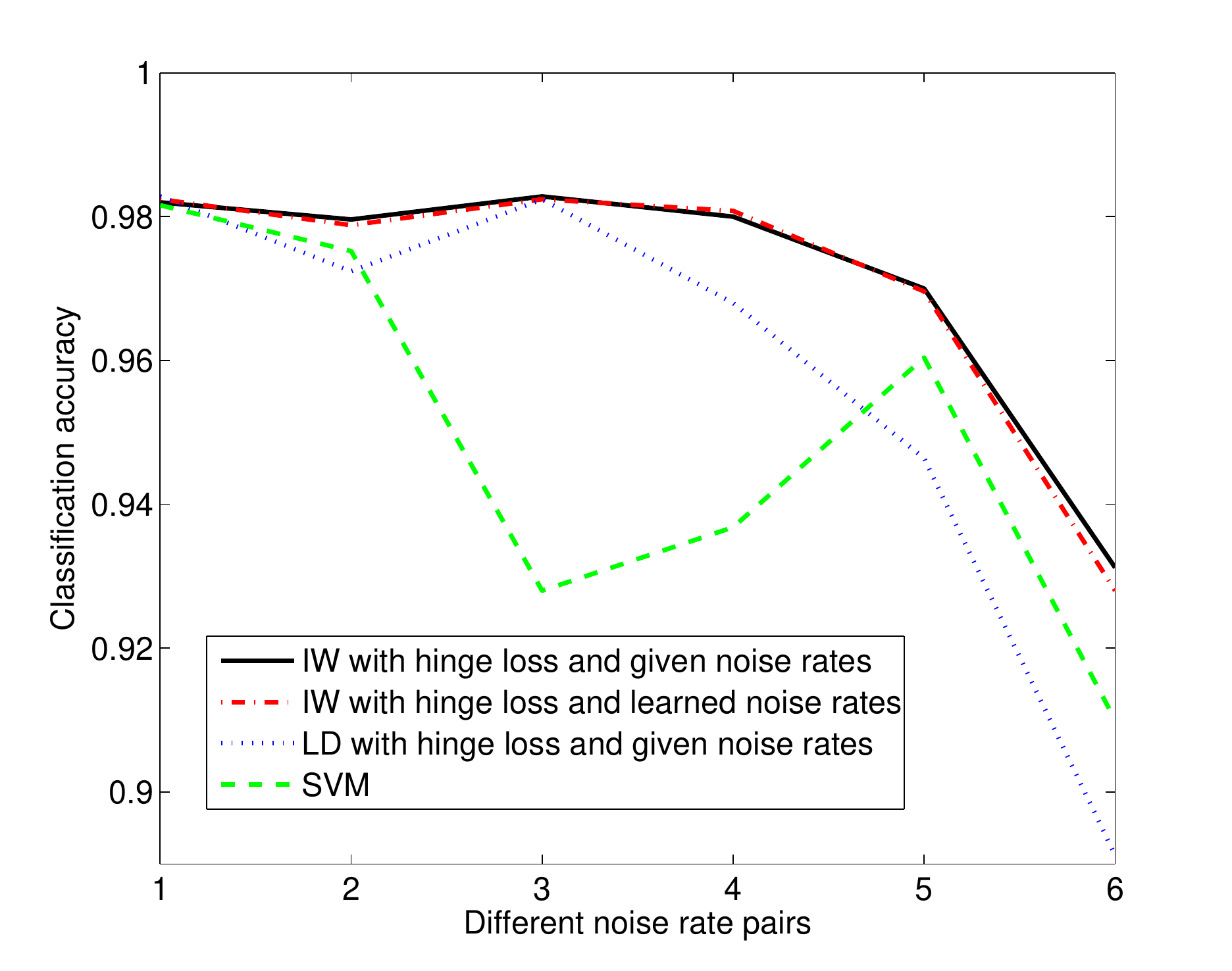}}\\
  \end{center}
  \caption{Accuracy comparison of classification algorithms on synthetic data (m=2, n=1000). The six different noise rate pairs ($\rho_{+1},\rho_{-1}$) are: (0.1, 0.1), (0.2, 0.2), (0.3, 0.1), (0.1, 0.3), (0.3, 0.3), and (0.4, 0.4). UB method employing hinge loss was not implemented due to non-convexity.}\label{syntheticExp}
\end{figure*}

\begin{table*}[!t]
\renewcommand{\arraystretch}{1.0}
\caption{Estimating the Noise Rates (Means and Standard Deviations) on Synthetic Data}\label{syntheticration}
\centering 
\resizebox{17cm}{!}{\small
\begin{tabular}[t]{|c|c|c|c|c|}
\hline
Dataset (m, n) & True ($\rho_{+1},\rho_{-1}$) & Cross-validation & Scott's Method & Our Method  \\
\cline{1-5}
Synthetic & (0, 0.4) & \textbf{(0.036$\pm$0.032, 0.402$\pm$0.153)} & \textbf{(0.000$\pm$0.000, 0.382$\pm$0.032)} & \textbf{(0.019$\pm$0.001, 0.422$\pm$0.019)} \\
\cline{2-5}
dataset & (0.1, 0.3) & (0.322$\pm$0.163, 0.352$\pm$0.158) & \textbf{(0.089$\pm$0.028, 0.297$\pm$0.032)} & \textbf{(0.125$\pm$0.023, 0.325$\pm$0.023)} \\
\cline{2-5}
(2, 1000) & (0.2, 0.2) & (0.424$\pm$0.098, 0.352$\pm$0.171) & \textbf{(0.176$\pm$0.035, 0.203$\pm$0.036)} & \textbf{(0.213$\pm$0.031, 0.230$\pm$0.023)} \\
\hline
Synthetic & (0, 0.4) & (0.445$\pm$0.016, 0.235$\pm$0.024) & \textbf{(0.000$\pm$0.000, 0.320$\pm$0.049)} & \textbf{(0.074$\pm$0.021, 0.458$\pm$0.112)} \\
\cline{2-5}
dataset & (0.1, 0.3) & (0.440$\pm$0.021, 0.310$\pm$0.091) & \textbf{(0.078$\pm$0.022, 0.255$\pm$0.047)} & \textbf{(0.140$\pm$0.022, 0.328$\pm$0.090)} \\
\cline{2-5}
(20, 1000) & (0.2, 0.2) & (0.425$\pm$0.043, 0.445$\pm$0.016) & \textbf{(0.159$\pm$0.022, 0.168$\pm$0.037)} & \textbf{(0.214$\pm$0.064, 0.226$\pm$0.023)} \\
\hline
\end{tabular}}
\label{complexity}
\end{table*}

Since $r(x)$ is usually modeled by linear or non-linear functions, we can assume that $r(x)$ has the range $[a,b]$ for all observations.  Using the Rademacher method again, for any $\delta>0$, with probability at least $1-\delta$, we have
\begin{eqnarray}\label{bregman}
&\text{BD}_f(r^*\|\hat{r})&\leq2\sup_r|\hat{\text{BD}}_f(r^*\|r)-\text{BD}_f(r^*\|r)|\nonumber\\
&&\leq 2\mathfrak{R}_\text{BDR}+C\sqrt{\frac{\log(1/\delta)}{2n}},
\end{eqnarray}
where $\mathfrak{R}_\text{BDR}$ defined in (\ref{comr})) is the Rademacher complexity induced by estimating the density ratio exploiting Bregman divergence and $C$ is a constant. The convergence rate of $\mathfrak{R}_{\text{BDR}}$ can be proven to be as fast as the order $\mathcal{O}(\sqrt{1/\min(n_+,n_-)})$, where $n_+$ and $n_-$ denote the number of positive labels and negative labels of the noisy sample, respectively. So, the ratio matching approach exploiting Bregman divergence is consistent to the optimal approximation in the hypothesis class.

\section{Experiments}\label{exp}

\begin{table*}
\centering \caption{Estimating the Noise Rates (Means and Standard Deviations) on UCI Benchmarks}\label{noiseUCI}
\renewcommand{\arraystretch}{1.0}
\resizebox{17cm}{!}{\small
\begin{tabular}[t]{|c|c|c|c|c|}
\hline
Dataset (m, n) & True ($\rho_{+1},\rho_{-1}$) & Cross-validation & Scott's Method & Our Method  \\
\cline{1-5}
 & (0, 0.4) & (0.186$\pm$0.150, 0.050$\pm$0.047) & (0.073$\pm$0.043, 0.225$\pm$0.080) & \textbf{(0.027$\pm$0.013, 0.364$\pm$0.078)} \\
\cline{2-5}
Heart & (0.1, 0.3) & \textbf{(0.134$\pm$0.102, 0.220$\pm$0.168)} & \textbf{(0.089$\pm$0.031, 0.231$\pm$0.116)} & \textbf{(0.070$\pm$0.020, 0.272$\pm$0.070)} \\
\cline{2-5}
(13, 270) & (0.2, 0.2) & \textbf{(0.110$\pm$0.109, 0.260$\pm$0.172)} & \textbf{(0.105$\pm$0.055, 0.179$\pm$0.047)} & \textbf{(0.121$\pm$0.063, 0.131$\pm$0.065)} \\
\hline
 & (0, 0.4) & (0.288$\pm$0.123, 0.206$\pm$0.118) & (0.122$\pm$0.036, 0.310$\pm$0.050) & \textbf{(0.026$\pm$0.012, 0.402$\pm$0.062)} \\
\cline{2-5}
Diabetes & (0.1, 0.3) & (0.154$\pm$0.099, 0.162$\pm$0.087) & (0.254$\pm$0.045, 0.249$\pm$0.057) & \textbf{(0.096$\pm$0.056, 0.304$\pm$0.060)} \\
\cline{2-5}
(8, 768) & (0.2, 0.2) & \textbf{(0.138$\pm$0.103, 0.168$\pm$0.131)} & (0.361$\pm$0.127, 0.185$\pm$0.053) & \textbf{(0.135$\pm$0.036, 0.215$\pm$0.074)} \\
\hline
\end{tabular}}
\label{complexity}
\end{table*}

\begin{table*}[t]
\centering \caption{Means and Standard Deviations (Percentage) of Classification Accuracies of all Kernel Hinge-loss-based Methods on UCI benchmarks}\label{realworld1}
\renewcommand{\arraystretch}{0.90}
\resizebox{14cm}{!}{\small
\begin{tabular}[t]{|c|c|c|c|c|c|c|}
\hline
{Bechmark dataset } & Noise rate & & & & & \\
{($m, n_+, n_-$)} & ($\rho_{+1}$, $\rho_{-1}$) & $\ell_{\text{hinge}}$ &LD$\ell_{\text{hinge}}$  & StPMKL & eIW$\ell_{\text{hinge}}$  & IW$\ell_{\text{hinge}}$ \\
\hline
                &  (0.2, 0.2)  &  64.24$\pm$6.59  &  65.76$\pm$9.29  &  \textbf{71.36$\pm$5.95}  &  69.39$\pm$5.91  &  \textbf{71.06$\pm$4.13} \\
Breast cancer   &  (0.3, 0.1)  &  67.12$\pm$8.69  &  70.61$\pm$5.16  &  \textbf{71.82$\pm$5.21}  &  68.79$\pm$7.57  &  \textbf{72.73$\pm$5.15} \\
(9, 77, 186)    &  (0.4, 0.4)  &  57.88$\pm$5.52  &  54.18$\pm$11.84 &  \textbf{67.12$\pm$8.72}  &  65.30$\pm$7.64  &  \textbf{68.79$\pm$8.09} \\
\hline
                &  (0.2, 0.2)  &  71.56$\pm$4.20  &  71.77$\pm$4.51  &  65.00$\pm$2.50  &  \textbf{73.02$\pm$3.09}  &  \textbf{72.92$\pm$3.53} \\
Diabetes        &  (0.3, 0.1)  &  \textbf{73.59$\pm$2.63}  &  73.23$\pm$2.37  &  66.46$\pm$2.75  &  \textbf{74.27$\pm$2.77}  &  71.46$\pm$3.04 \\
(8, 268, 500)   &  (0.4, 0.4)  &  66.77$\pm$2.37  &  66.25$\pm$4.31  &  \textbf{73.18$\pm$4.01}  &  \textbf{71.98$\pm$2.50}  &  71.77$\pm$3.38 \\
\hline
                &  (0.2, 0.2)  &  67.20$\pm$3.55  &  \textbf{68.68$\pm$2.84}  &  \textbf{69.80$\pm$2.23}  &  67.20$\pm$3.45  &  68.08$\pm$2.85 \\
German          &  (0.3, 0.1)  &  68.56$\pm$2.62  &  \textbf{70.84$\pm$2.87}  &  67.24$\pm$1.78  &  68.76$\pm$2.29  &  \textbf{69.56$\pm$2.37} \\
(20, 300, 700)  &  (0.4, 0.4)  &  62.32$\pm$2.81  &  62.04$\pm$5.90  &  \textbf{71.96$\pm$3.41}  &  \textbf{63.36$\pm$2.83}  &  63.04$\pm$2.89 \\
\hline
                &  (0.2, 0.2)  &  67.21$\pm$5.33  &  \textbf{70.15$\pm$6.20}  &  \textbf{77.21$\pm$9.88}  &  68.82$\pm$5.62  &  68.82$\pm$5.08 \\
Heart           &  (0.3, 0.1)  &  70.59$\pm$8.05  &  \textbf{72.21$\pm$8.16}  &  54.71$\pm$7.96  &  70.74$\pm$8.42  &  \textbf{72.06$\pm$6.69} \\
(13, 120, 150)  &  (0.4, 0.4)  &  68.68$\pm$6.12  &  67.94$\pm$13.28 &  59.12$\pm$13.30 &  \textbf{70.29$\pm$5.62}  &  \textbf{69.71$\pm$6.09} \\
\hline
                &  (0.2, 0.2)  &  \textbf{92.80$\pm$1.19}  &  92.16$\pm$0.95  &  73.35$\pm$1.80  &  \textbf{92.82$\pm$1.14}  &  92.49$\pm$0.93 \\
Image           &  (0.3, 0.1)  &  91.30$\pm$1.99  &  \textbf{91.74$\pm$2.29}  &  58.89$\pm$6.72  &  91.02$\pm$1.70  &  \textbf{92.07$\pm$2.27} \\
(18, 1188, 898) &  (0.4, 0.4)  &  \textbf{91.97$\pm$3.18}  &  90.98$\pm$1.49  &  57.24$\pm$2.83  &  \textbf{92.13$\pm$1.33}  &  89.37$\pm$3.45 \\
\hline
                &  (0.2, 0.2)  &  \textbf{89.81$\pm$3.18}  &  \textbf{90.74$\pm$3.38}  &  70.93$\pm$3.50  &  88.52$\pm$3.58  &  87.41$\pm$4.43 \\
Thyroid         &  (0.3, 0.1)  &  \textbf{87.22$\pm$6.95}  &  \textbf{90.93$\pm$5.89}  &  69.81$\pm$6.93  &  85.19$\pm$7.81  &  81.85$\pm$6.58 \\
(5, 65, 150)    &  (0.4, 0.4)  &  \textbf{91.85$\pm$4.02}  &  88.52$\pm$13.26 &  82.22$\pm$12.60 &  91.76$\pm$2.66  &  \textbf{93.15$\pm$3.50} \\
\hline
Average         &              &  75.04           &   75.44          &  68.19           &  \textbf{76.29}           &  \textbf{76.48}          \\
\hline
\end{tabular}}
\label{complexity}
\end{table*}

\begin{table*}[t]
\centering \caption{Means and Standard Deviations (Percentage) of Classification Accuracies of all Kernel Logistic-loss-based Methods on UCI benchmarks}\label{realworld2}
\renewcommand{\arraystretch}{0.90}
\resizebox{14cm}{!}{\small
\begin{tabular}[t]{|c|c|c|c|c|c|c|}
\hline
{Bechmark dataset } & Noise rate && &  &  & \\
{($m, n_+, n_-$)} & ($\rho_{+1}$, $\rho_{-1}$) & $\ell_\text{log}$ &LD$\ell_\text{log}$   &UB$\ell_\text{log}$  &eIW$\ell_\text{log}$   &IW$\ell_\text{log}$   \\
\hline
               &  (0.2, 0.2)  &  73.48$\pm$5.16  &  \textbf{73.48$\pm$4.47}  &  72.73$\pm$4.46  &  72.88$\pm$6.04  &  \textbf{73.94$\pm$4.33}  \\
Breast cancer  &  (0.3, 0.1)  &  \textbf{73.33$\pm$3.86}  &  70.91$\pm$5.09  &  71.67$\pm$5.49  &  71.36$\pm$5.41  &  \textbf{71.97$\pm$5.99}  \\
(9, 77, 186)   &  (0.4, 0.4)  &  66.36$\pm$8.41  &  \textbf{67.73$\pm$11.50} &  67.09$\pm$8.24  &  \textbf{71.76$\pm$6.89}  &  65.61$\pm$7.69  \\
\hline
               &  (0.2, 0.2)  &  \textbf{74.43$\pm$2.67}  &  72.24$\pm$2.78  &  72.92$\pm$2.95  &  \textbf{73.70$\pm$2.49}  &  72.45$\pm$2.74  \\
Diabetes       &  (0.3, 0.1)  &  \textbf{73.54$\pm$2.98}  &  73.12$\pm$4.29  &  72.34$\pm$4.71  &  \textbf{73.70$\pm$2.47}  &  73.33$\pm$3.62  \\
(8, 268, 500)  &  (0.4, 0.4)  &  70.21$\pm$4.56  &  71.04$\pm$5.10  &  \textbf{71.30$\pm$4.56}  &  \textbf{73.85$\pm$3.50}  &  70.83$\pm$3.67  \\
\hline
               &  (0.2, 0.2)  &  69.28$\pm$2.20  &  68.80$\pm$2.66  &  \textbf{69.52$\pm$2.11}  &  \textbf{69.72$\pm$2.02}  &  69.00$\pm$2.76 \\
German         &  (0.3, 0.1)  &  \textbf{67.36$\pm$1.91}  &  67.20$\pm$2.17  &  67.28$\pm$1.89  &  \textbf{67.36$\pm$1.92}  &  67.32$\pm$2.04 \\
(20, 300, 700) &  (0.4, 0.4)  &  60.60$\pm$7.13  &  60.36$\pm$9.15  &  \textbf{65.16$\pm$6.66}  &  \textbf{64.96$\pm$6.19}  &  64.56$\pm$6.95 \\
\hline
               &  (0.2, 0.2)  &  \textbf{82.21$\pm$5.61}  &  80.29$\pm$7.44  &  \textbf{82.94$\pm$5.01}  &  81.32$\pm$10.36 &  81.91$\pm$4.44 \\
Heart          &  (0.3, 0.1)  &  69.41$\pm$9.37  &  \textbf{77.06$\pm$8.80}  &  75.88$\pm$9.02  &  75.44$\pm$9.33  &  \textbf{76.91$\pm$7.84} \\
(13, 120, 150) &  (0.4, 0.4)  &  69.56$\pm$10.17 &  \textbf{78.24$\pm$5.62}  &  76.18$\pm$7.03  &  \textbf{78.38$\pm$9.40}  &  77.50$\pm$7.04 \\
\hline
               &  (0.2, 0.2)  &  \textbf{62.84$\pm$3.02}  &  59.85$\pm$7.36  &  \textbf{65.84$\pm$3.70}  &  62.16$\pm$4.68  &  61.72$\pm$4.88 \\
Image          &  (0.3, 0.1)  &  \textbf{58.56$\pm$2.72}  &  57.47$\pm$1.82  &  56.15$\pm$1.90  &  \textbf{58.91$\pm$3.02}  &  58.26$\pm$2.69 \\
(18, 1188, 898)&  (0.4, 0.4)  &  60.48$\pm$7.60  &  63.72$\pm$4.36  &  \textbf{65.27$\pm$3.95}  &  \textbf{64.69$\pm$5.60}  &  62.26$\pm$4.07 \\
\hline
               &  (0.2, 0.2)  &  89.07$\pm$4.40  &  \textbf{90.93$\pm$3.08}  &  90.37$\pm$3.47  &  86.11$\pm$6.37  &  \textbf{92.41$\pm$2.82} \\
Thyroid        &  (0.3, 0.1)  &  84.26$\pm$4.12  &  \textbf{88.89$\pm$4.78}  &  85.04$\pm$6.36  &  82.59$\pm$4.38  &  \textbf{87.96$\pm$5.26} \\
(5, 65, 150)   &  (0.4, 0.4)  &  86.48$\pm$6.88  &  87.04$\pm$9.28  &  86.30$\pm$9.16  &  \textbf{88.33$\pm$6.11}  &  \textbf{88.70$\pm$4.23} \\
\hline
Average         &              &  71.75           &   72.69          &  73.00           &  \textbf{73.13}           &  \textbf{73.23}          \\
\hline
\end{tabular}}
\label{complexity}
\end{table*}

We next conducted experiments on synthetic and real data to illustrate the performance of the proposed approaches. Each dataset was randomly split 10 times, 75\% for training and 25\% for testing, and then the labels of the training sample flipped according to given noise rates $\rho_{+1}$ and $\rho_{-1}$. The mean accuracies of the 10 datasets are presented. 

To show the efficiency of our method for estimating the noise rates, we employed two baselines for comparison: the simple cross-validation method used in \cite{natarajan2013learning} and Scott's method \cite{scott2015a} for estimating the inversed noise rates. Note that Scott's method can not be exploited to estimate the noise rates unless the knowledge $P_D(\pm 1)$ is given, which is often unknown in practice. To make the comparison, we assumed that the knowledge is known.

For the task of classification with noisy labels, the unbiased estimator (UB$\ell$) and label-dependent costs (LD$\ell$) models, developed by Natarajan et al. \cite{natarajan2013learning}, and empirically shown to be competitive with, and perform better more often than, three of state-of-the-art robust methods (random projection classifier \cite{stempfel2007learning}, NHERD \cite{crammer2010learning}, and the perceptron algorithm with margin \cite{khardon2007noise}) for dealing with asymmetric RCN, were chosen as baselines for comparison\footnote{Note that comparisons in our paper and those in \cite{natarajan2013learning} are implemented on the same standard UCI classification datasets provided by Gunnar R\"atsch: \url{http://theoval.cmp.uea.ac.uk/matlab}.}. We denote our importance reweighting method by IW$\ell$ that estimates the conditional distribution $P_{D_{\rho}}(\hat{Y}|X)$ by employing the KLIEP method and the noise rates by using the cross-validation method, and denote our importance reweighting method by eIW$\ell$ that exploits the KLIEP method to estimate the conditional distribution $P_{D_{\rho}}(\hat{Y}|X)$ and the noise rates jointly. Three-fold cross-validation\footnote{To achieve high performance for UB$\ell$, LD$\ell$ and IW$\ell$, the optimal noise rates were chosen by the criterion that the classification accuracy rate, instead of the weighted objective function, is minimized on the validation set.} was used to tune the noise rates on the training sets when needed.

\subsection{Synthetic Data}
We first tested the performance of noise rate estimation on the synthetic dataset, where the data were uniformly distributed from the interval $[0,1]$ and then linearly separated into two classes such that $P(Y=+1)=P(Y=-1)$. We used kernel density and density ratio estimation methods to estimate the noise rates on 2-dimensional and 20-dimensional synthetic data, respectively. The kernel width for kernel density estimation method was chosen as the standard deviation, and the density ratio was estimated using the KLIEP method \cite{sugiyama2010density}. The performances of the different methods are shown in Table \ref{syntheticration}, with entries having errors less than 0.1 shown in bold. Table \ref{syntheticration} shows that our and Scott's methods for estimating the noise rates is far more accurate than the simple cross-validation method and that our method is comparable with that of Scott on the synthetic datasets.

\begin{table*}[t]
\centering \caption{Means and Standard Deviations (Percentage) of Classification Accuracies of all Linear Hinge-loss-based Methods on UCI benchmarks}\label{realworld3}
\renewcommand{\arraystretch}{0.90}
\resizebox{12cm}{!}{\small
\begin{tabular}[t]{|c|c|c|c|c|c|}
\hline
{Bechmark dataset } & Noise rate && &  &  \\
{($m, n_+, n_-$)} & ($\rho_{+1}$, $\rho_{-1}$) & $\ell_\text{hinge}$  &LD$\ell_\text{hinge}$  &eIW$\ell_\text{hinge}$   &IW$\ell_\text{hinge}$   \\
\hline
               &  (0.2, 0.2)  &  \textbf{71.82$\pm$3.93}  &  69.24$\pm$5.94  &  \textbf{71.36$\pm$4.65}  &  70.30$\pm$4.58        \\
Breast cancer  &  (0.3, 0.1)  &  71.52$\pm$3.56  &  71.36$\pm$3.67  &  \textbf{72.42$\pm$3.83}  &  \textbf{72.42$\pm$3.90}        \\
(9, 77, 186)   &  (0.4, 0.4)  &  66.36$\pm$7.55  &  63.64$\pm$13.59 &  \textbf{71.67$\pm$4.58}  &  \textbf{67.88$\pm$4.83}        \\
\hline
               &  (0.2, 0.2)  &  \textbf{76.88$\pm$1.89}  &  75.68$\pm$2.37  &  \textbf{75.99$\pm$6.55}  &  75.63$\pm$2.86        \\
Diabetes       &  (0.3, 0.1)  &  68.39$\pm$5.31  &  \textbf{73.59$\pm$5.19}  &  70.52$\pm$4.90  &  \textbf{74.22$\pm$4.99}        \\
(8, 268, 500)  &  (0.4, 0.4)  &  73.54$\pm$4.65  &  \textbf{75.05$\pm$5.43}  &  \textbf{76.41$\pm$2.38}  &  74.53$\pm$4.96        \\
\hline
               &  (0.2, 0.2)  &  71.20$\pm$2.80  &  71.08$\pm$2.87  &  \textbf{71.56$\pm$2.62}  &  \textbf{71.76$\pm$1.90}       \\
German         &  (0.3, 0.1)  &  67.16$\pm$1.91  &  \textbf{67.16$\pm$1.78}  &  \textbf{67.24$\pm$1.78}  &  67.16$\pm$1.91      \\
(20, 300, 700) &  (0.4, 0.4)  &  70.24$\pm$3.99  &  70.96$\pm$3.21  &  \textbf{71.08$\pm$2.41}  &  \textbf{72.56$\pm$3.44}       \\
\hline
               &  (0.2, 0.2)  &  78.82$\pm$5.20  &  77.35$\pm$5.29  &  \textbf{81.18$\pm$5.49}  &  \textbf{78.97$\pm$6.51}       \\
Heart          &  (0.3, 0.1)  &  \textbf{75.88$\pm$8.06}  &  74.12$\pm$9.49  &  \textbf{79.26$\pm$4.46}  &  75.59$\pm$7.08       \\
(13, 120, 150) &  (0.4, 0.4)  &  75.88$\pm$4.96  &  74.71$\pm$10.18 &  \textbf{78.97$\pm$5.05}  &  \textbf{78.38$\pm$6.76}       \\
\hline
               &  (0.2, 0.2)  &  79.62$\pm$2.55  &  \textbf{82.30$\pm$2.03}  &  79.69$\pm$2.66  &  \textbf{82.55$\pm$2.21}       \\
Image          &  (0.3, 0.1)  &  76.13$\pm$4.38  &  \textbf{82.09$\pm$1.74}  &  75.70$\pm$4.21  &  \textbf{82.78$\pm$1.30}       \\
(18, 1188, 898)&  (0.4, 0.4)  &  73.74$\pm$2.00  &  \textbf{81.84$\pm$2.53}  &  73.83$\pm$1.92  &  \textbf{80.21$\pm$4.20}       \\
\hline
               &  (0.2, 0.2)  &  85.00$\pm$6.32  &  \textbf{87.78$\pm$4.11}  &  82.59$\pm$7.67  &  \textbf{88.33$\pm$2.90}       \\
Thyroid        &  (0.3, 0.1)  &  82.22$\pm$5.03  &  \textbf{85.19$\pm$6.23}  &  77.59$\pm$6.95  &  \textbf{84.26$\pm$4.96}       \\
(5, 65, 150)   &  (0.4, 0.4)  &  \textbf{86.11$\pm$5.18}  &  \textbf{85.93$\pm$4.55}  &  85.56$\pm$7.40  &  85.00$\pm$7.06       \\
\hline
Average         &              &  75.03           &   \textbf{76.06}          &  75.72         &  \textbf{76.81}               \\
\hline
\end{tabular}}
\label{complexity}
\end{table*}

\begin{table*}[t]
\centering \caption{Means and Standard Deviations (Percentage) of Classification Accuracies of all Linear Logistic-loss-based Methods on UCI benchmarks}\label{realworld4}
\renewcommand{\arraystretch}{0.90}
\resizebox{14cm}{!}{\small
\begin{tabular}[t]{|c|c|c|c|c|c|c|}
\hline
{Bechmark dataset } & Noise rate && &  &  & \\
{($m, n_+, n_-$)} & ($\rho_{+1}$, $\rho_{-1}$) & $\ell_\text{log}$ &LD$\ell_\text{log}$  &UB$\ell_\text{log}$   &eIW$\ell_\text{log}$   &IW$\ell_\text{log}$   \\
\hline
               &  (0.2, 0.2)  &  70.00$\pm$5.88  &  71.52$\pm$4.83  &  \textbf{71.82$\pm$5.31}  &  \textbf{73.48$\pm$4.91}  &  71.52$\pm$4.51    \\
Breast cancer  &  (0.3, 0.1)  &  \textbf{72.42$\pm$3.96}  &  70.76$\pm$4.63  &  \textbf{72.73$\pm$4.23}  &  71.06$\pm$4.49  &  71.06$\pm$4.76    \\
(9, 77, 186)   &  (0.4, 0.4)  &  63.33$\pm$6.42  &  60.30$\pm$16.24 &  61.52$\pm$16.12 &  \textbf{66.97$\pm$6.95}  &  \textbf{65.61$\pm$12.00}    \\
\hline
               &  (0.2, 0.2)  &  \textbf{77.14$\pm$1.84}  &  \textbf{77.29$\pm$2.00}  &  76.51$\pm$2.22  &  74.79$\pm$2.52  &  76.72$\pm$2.00    \\
Diabetes       &  (0.3, 0.1)  &  74.48$\pm$2.33  &  \textbf{74.79$\pm$2.96}  &  \textbf{74.64$\pm$3.30}  &  74.48$\pm$3.44  &  74.37$\pm$3.49    \\
(8, 268, 500)  &  (0.4, 0.4)  &  71.20$\pm$3.17  &  72.86$\pm$5.29  &  71.93$\pm$4.89  &  \textbf{77.03$\pm$3.39}  &  \textbf{74.06$\pm$5.01}    \\
\hline
               &  (0.2, 0.2)  &  \textbf{72.80$\pm$1.73}  &  72.68$\pm$1.93  &  \textbf{72.92$\pm$2.14}  &  71.56$\pm$2.95  &  72.64$\pm$1.28   \\
German         &  (0.3, 0.1)  &  \textbf{70.88$\pm$2.40}  &  68.76$\pm$1.96  &  70.04$\pm$2.03  &  \textbf{71.76$\pm$3.13}  &  69.08$\pm$2.52   \\
(20, 300, 700) &  (0.4, 0.4)  &  70.60$\pm$3.23  &  70.88$\pm$5.23  &  \textbf{71.56$\pm$3.96}  &  \textbf{71.20$\pm$3.70}  &  70.84$\pm$5.20   \\
\hline
               &  (0.2, 0.2)  &  77.50$\pm$3.80  &  76.91$\pm$6.01  &  77.79$\pm$5.74  &  \textbf{78.97$\pm$6.61}  &  \textbf{79.85$\pm$5.33}   \\
Heart          &  (0.3, 0.1)  &  74.85$\pm$7.02  &  74.26$\pm$6.36  &  74.41$\pm$9.56  &  \textbf{78.68$\pm$5.85}  &  \textbf{75.00$\pm$7.17}   \\
(13, 120, 150) &  (0.4, 0.4)  &  74.71$\pm$5.27  &  73.82$\pm$7.49  &  \textbf{77.21$\pm$5.60}  &  \textbf{77.79$\pm$6.81}  &  72.50$\pm$6.09   \\
\hline
               &  (0.2, 0.2)  &  \textbf{82.22$\pm$2.31}  &  82.05$\pm$2.21  &  81.99$\pm$2.76  &  82.18$\pm$2.47  &  \textbf{82.43$\pm$2.23}   \\
Image          &  (0.3, 0.1)  &  73.24$\pm$4.28  &  80.29$\pm$2.67  &  \textbf{80.63$\pm$2.88}  &  72.76$\pm$3.92  &  \textbf{81.19$\pm$2.35}   \\
(18, 1188, 898)&  (0.4, 0.4)  &  77.59$\pm$1.79  &  81.32$\pm$2.03  &  \textbf{82.09$\pm$2.27}  &  77.93$\pm$1.85  &  \textbf{81.59$\pm$2.52}   \\
\hline
               &  (0.2, 0.2)  &  \textbf{85.37$\pm$5.20}  &  85.00$\pm$4.14  &  85.00$\pm$5.12  &  84.07$\pm$6.66  &  \textbf{86.11$\pm$3.52}   \\
Thyroid        &  (0.3, 0.1)  &  82.22$\pm$4.88  &  \textbf{84.07$\pm$5.74}  &  82.22$\pm$5.60  &  81.30$\pm$4.14  &  \textbf{84.63$\pm$4.46}  \\
(5, 65, 150)   &  (0.4, 0.4)  &  80.37$\pm$6.99  &  85.37$\pm$4.66  &  84.26$\pm$5.67  &  \textbf{86.30$\pm$6.43}  &  \textbf{86.48$\pm$5.24}   \\
\hline
Average         &              &  75.05           &   75.72          &  76.07           &  \textbf{76.24}           &  \textbf{76.43}          \\
\hline
\end{tabular}}
\label{complexity}
\end{table*}

We next tested the performance of IW$\ell$, UB$\ell$, and LD$\ell$ on the synthetic data. For fair comparison, we used the true noise rates for each model so that there was no tuning parameter for all the methods. Even though our method still needs to estimate the conditional probability $P_{D_\rho}(y|x)$, Fig. \ref{syntheticExp} shows that our method is more effective than the baselines when tested on these synthetic data. The empirical results also show that the logistic classification and SVM perform very bad when the noise rates are asymmetric.

\subsection{Comparison on UCI Benchmarks}
The comparison of noise rate estimation on two UCI datasets are shown in Table \ref{noiseUCI}, with entries having error less than 0.1 shown in bold. The results show that the cross-validation method can estimate some noise rates with errors less than 0.1. However, this method produced large standard deviations. We employed the KLIEP method to estimate the noise rates. Table \ref{noiseUCI} illustrates that our method is more accurate than the baselines and has errors and standard deviations less than 0.1. These errors in our method occurred for two reasons: the first is that no example has very small $P_D(Y|X)$ in the Euclidian space (according to the theory part, in this case, both Scott's and our methods cannot perform well), and the second is that the KLIEP method for estimating the conditional distribution is inaccurate due to the difficulty in choosing kernel width. However, when using the estimated noise rates, the performance of our method (eIW$\ell$) for classification with noisy labels performs better more often than the baselines (see Tables \ref{realworld1}, \ref{realworld2}, \ref{realworld3} and \ref{realworld4}). Our noise rate estimation method is also valuable in the sense that some methods can benefit when the noise rates are approximately known \cite{natarajan2013learning}.

To further demonstrate the efficiency of our importance reweighting method, we also tested multiple kernel learning from noisy labels by stochastic programming (StPMKL) \cite{yang2012multiple} as a baseline on six UCI datasets, where all single kernel learning methods used Gaussian kernel with width 1 and StPMKL used Gaussian kernels with 10 different widths $\{2^{-3}, 2^{-2},\ldots,2^{6}\}$. The performances of the methods for different noise rates are shown in Tables \ref{realworld1}, \ref{realworld2}, \ref{realworld3} and \ref{realworld4}, separately, with the two highest values in each row shown in bold. The results validate that our methods eIW$\ell$, which employ the estimated noise rates and run fast, perform better more often than the baselines in all datasets and that our methods IW$\ell$ have average performances better than those of all the baselines. 
While our methods IW have average performances better than those of all the baselines, the methods as well as the baselines LD and UB need to learn the noise rates via cross-validation, which is time-consuming.
Note that the kernel logistic-loss-based average performances in Tables \ref{realworld2} are lower than those in Tables \ref{realworld1}, \ref{realworld3} and \ref{realworld4} because the kernel logistic-loss-based methods have low accuracies on the Image dataset.

According to Tables \ref{realworld1}, \ref{realworld2}, \ref{realworld3} and \ref{realworld4}, the method IW$\ell_\text{hinge}$, which has the highest average performances, is preferred in practice. However, the method eIW$\ell_\text{hinge}$, which is time efficient and is competitive with all the baselines on all the datasets, should be sometimes preferred because of its low training time complexity.

\section{Conclusions and Future Work}\label{con}
In this paper, we presented an importance reweighting framework for classification in the presence of label noise. Theoretical analyses were provided to assure that the learned classifier will converge to the optimal one for the noise-free sample. Empirical studies on synthetic and real-world datasets verified the effectiveness and robustness of our proposed learning framework. We also provided a method to estimate the noise rates.

All our proposed methods crucially depend on the accuracy of the estimation of the conditional distribution $P_{D_{\rho}}(\hat{Y}|X)$. In future work, we need to consider how to accurately learn the conditional probability distribution for the noisy sample. 



%

\appendices

\section*{Acknowledgment}
We greatly thank the handling Associate Editor and all anonymous reviewers for their valuable comments. The work was supported in part by Australian Research Council Projects DP-140102164 and FT-130101457.

%
%
\ifCLASSOPTIONcaptionsoff
  \newpage
\fi

\bibliographystyle{ieeetr}
\bibliography{reference}

\begin{IEEEbiography}[{\includegraphics[width=1in,height=1.25in,clip,keepaspectratio]{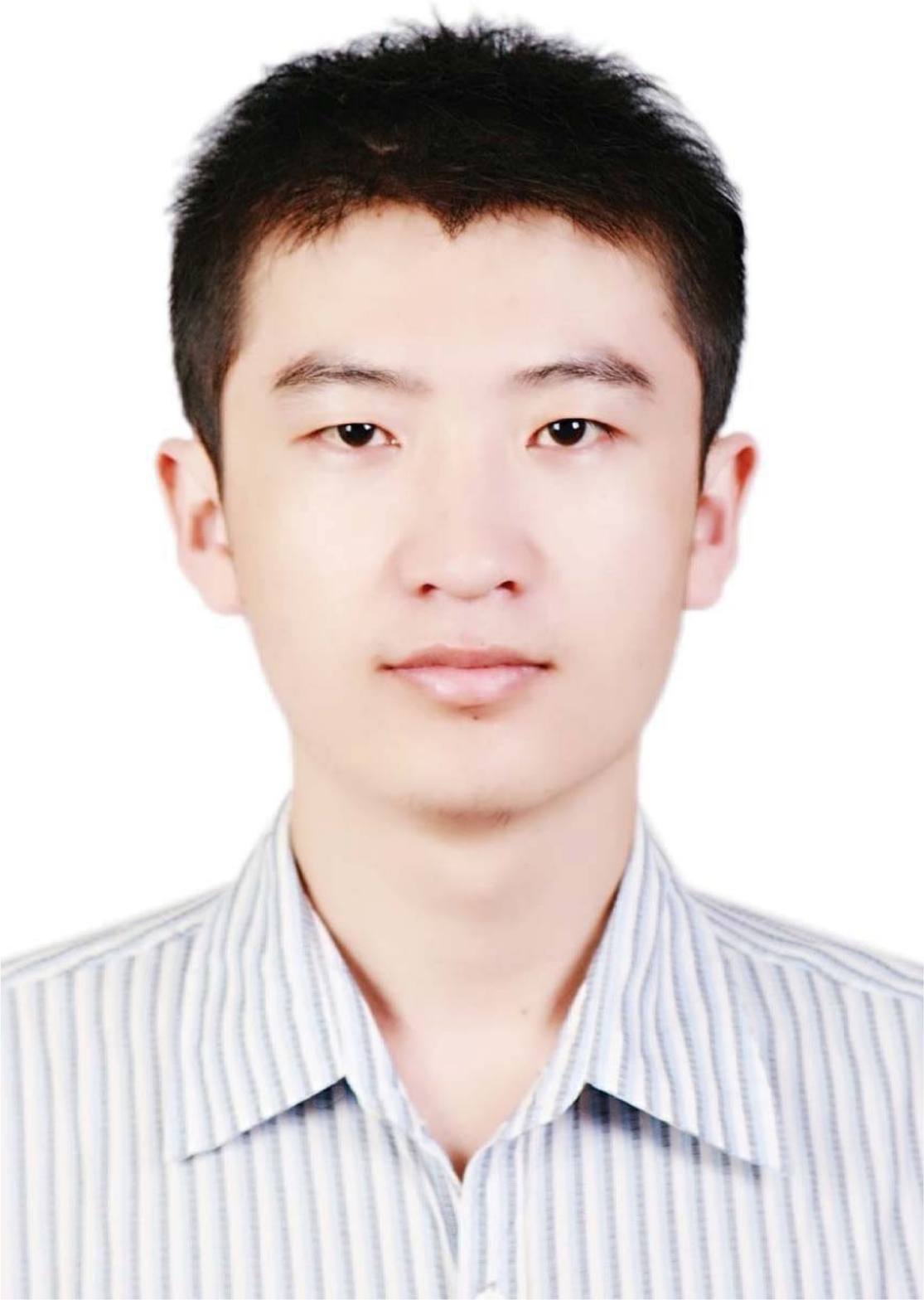}}]{Tongliang Liu}
received the B.E. degree in electronics engineering and information science from the University of Science and Technology of China, in 2012. He is currently pursuing the Ph.D. degree in computer science from the University of Technology, Sydney. He won the best paper award in the IEEE International Conference on Information Science and Technology 2014. 

His research interests include statistical learning theory,
computer vision, and optimization.

\end{IEEEbiography}

\begin{IEEEbiography}[{\includegraphics[width=1in,height=1.25in,clip,keepaspectratio]{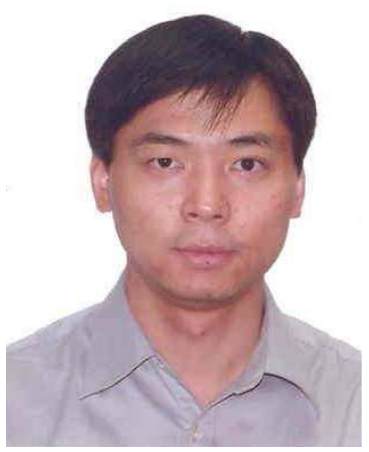}}]{Dacheng Tao}
(F'15) is Professor of Computer Science with the Centre for Quantum Computation $\&$ Intelligent Systems, and the Faculty of Engineering and Information Technology in the University of Technology, Sydney. He mainly applies statistics and mathematics to data analytics problems and his research interests spread across computer vision, data science, image processing, machine learning, and video surveillance. His research results have expounded in one monograph and 100+ publications at prestigious journals and prominent conferences, such as IEEE T-PAMI, T-NNLS, T-IP, JMLR, IJCV, NIPS, ICML, CVPR, ICCV, ECCV, AISTATS, ICDM; and ACM SIGKDD, with several best paper awards, such as the best theory/algorithm paper runner up award in IEEE ICDM’07, the best student paper award in IEEE ICDM’13, and the 2014 ICDM 10 Year Highest-Impact Paper Award.
\end{IEEEbiography}



\newpage

The following is the supplementary material to the main manuscript providing proofs of Theorem 3, Proposition 2 and the assertions in Remark 2.
\newpage

\begin{appendices}
\numberwithin{equation}{section}

\section{Proofs of Theorem 3 and Proposition 2}
\textbf{Proof of Theorem 3.}
If the hypothesis class for estimating the density ratio is set properly so that the approximation error is zero, the target density ratio $r^*(X)=P_{D_\rho}(X|Y)/P_{D_\rho}(X)$ will be included in the hypothesis class. The consistency of the ratio matching approach exploiting Bregman divergence (proven in Section 7.6 of the submission) guarantees that the target density ratio $r^*(X)$ can be learned when $n$ is sufficiently large. Using the proof method of Theorem 2, for any $\epsilon>0$, we can prove that
\begin{eqnarray*}
&&\lim_{n\rightarrow\infty}P(R[D,\hat{f}_{n,\hat{\beta}},\ell(\hat{f}_{n,\hat{\beta}}(X),Y)]\\
&&\ \ \ \ \ \ \ \ \ \ \ \ \ \ \ \ \ \ \ -R[D,f^*,\ell(f^*(X),Y)]> \epsilon)=0.
\end{eqnarray*}
\hfill$\blacksquare$

\textbf{Proof of Proposition 2.}
If we let $f(r)=(t-1)^2/2$, the Bregman divergence degenerates to the square distance as follows
$$\text{BD}_f(r^*\|r)=\text{SD}(r^*\|r)=\frac{1}{2}(r^*-r)^2.$$
According to (9), for any $\delta>0$, with probability at least $1-\delta$, we have
\begin{eqnarray*}
&\text{BD}(r^*\|\hat{r})&=\frac{1}{2}(r^*-\hat{r})^2\\
&&\leq2\sup_r|\hat{\text{BD}}(r^*\|r)-\text{BD}(r^*\|r)|\\
&&\leq 2\mathfrak{R}_{\text{BDR}}+C\sqrt{\frac{\log(1/\delta)}{2n}},
\end{eqnarray*}
where $\mathfrak{R}_{\text{BDR}}$ is defined in (8).
Thus, with probability at least $1-\delta$, the following holds
\begin{eqnarray}\label{dre1}
&&|r^*-\hat{r}|\leq\mathcal{O}\left(\sqrt{\mathfrak{R}_{\text{BDR}}+\sqrt{\frac{\log(1/\delta)}{n}}}\right).
\end{eqnarray}

The convergence rate of $\mathfrak{R}_{\text{BDR}}$ can be proven to be of order $\mathcal{O}(\sqrt{1/\min(n_+,n_-)})$, where $n_+$ and $n_-$ denote the number of positive labels and negative labels of the noisy sample, respectively.

Note that $P_{D_{\rho}}(\hat{Y}|X)=\frac{P_{D_{\rho}}(X|\hat{Y})P_{D_{\rho}}(\hat{Y})}{P_{D_{\rho}}(X)}$, and that we estimate $\frac{P_{D_{\rho}}(X|\hat{Y})}{P_{D_{\rho}}(X)}$ by employing the Bregman divergence based ratio matching method and $P_{D_{\rho}}(\hat{Y}=\pm1)$ by $\frac{1}{n}\sum_{i=1}^{n}1_{\{\hat{Y}=\pm1\}}$. 

Using Hoeffding's inequality, with probability at least $1-\delta$, we have 
\begin{eqnarray}\label{dre2}
\left|P_{D_{\rho}}(\hat{Y}=\pm1) -\hat{P}_{D_{\rho}}(\hat{Y}=\pm1)\right|\leq\sqrt{\frac{\log(1/\delta)}{2n}}.
\end{eqnarray}

Combining Equations (\ref{dre1}) and (\ref{dre2}), with probability at least $1-2\delta$, we have
\begin{eqnarray*}
&&\left|P_{D_{\rho}}(\hat{Y}|X)- \hat{P}_{D_{\rho}}(\hat{Y}|X)\right|\\ 
&&\leq\mathcal{O}\left(\sqrt{\frac{\log(1/\delta)}{n}}+\sqrt{\mathfrak{R}_{\text{SDR}}+\sqrt{\frac{\log(1/\delta)}{n}}}\right).\nonumber
\end{eqnarray*} 
Then, with probability at least $1-2\delta$, we have
\begin{eqnarray*}
&&{\beta}(X,\hat{Y})\\ 
&&=\frac{{P}_{D_\rho}(\hat{Y}|X)-\rho_{-\hat{Y}}}{(1-\rho_{-1}-\rho_{+1}){P}_{D_\rho}(\hat{Y}|X)}\\ 
&&=\frac{1-\frac{\rho_{-\hat{Y}}}{{P}_{D_\rho}(\hat{Y}|X)}}{1-\rho_{-1}-\rho_{+1}}\\ 
&&\leq \frac{1-\frac{\rho_{-\hat{Y}}}{\hat{P}_{D_\rho}(\hat{Y}|X)+\mathcal{O}\left(\sqrt{\frac{\log(1/\delta)}{n}}+\sqrt{\mathfrak{R}_{\text{SDR}}+\sqrt{\frac{\log(1/\delta)}{n}}}\right)}}{1-\rho_{-1}-\rho_{+1}}\\
\end{eqnarray*}
Let $\Delta(n)\triangleq\mathcal{O}\left(\sqrt{\frac{\log(1/\delta)}{n}}+\sqrt{\mathfrak{R}_{\text{SDR}}+\sqrt{\frac{\log(1/\delta)}{n}}}\right)$, we have
\begin{eqnarray*}
&&{\beta}(X,\hat{Y})\\ 
&&\leq\frac{\hat{P}_{D_\rho}(\hat{Y}|X)+\Delta(n)-\rho_{-\hat{Y}}}{(1-\rho_{-1}-\rho_{+1})(\hat{P}_{D_\rho}(\hat{Y}|X)+\Delta(n))}\\
&&\leq\frac{\hat{P}_{D_\rho}(\hat{Y}|X)+\Delta(n)-\rho_{-\hat{Y}}}{(1-\rho_{-1}-\rho_{+1})\hat{P}_{D_\rho}(\hat{Y}|X)}\\ 
&&\leq\frac{\hat{P}_{D_\rho}(\hat{Y}|X)-\rho_{-\hat{Y}}}{(1-\rho_{-1}-\rho_{+1})\hat{P}_{D_\rho}(\hat{Y}|X)}\\
&&\ \ \ +\mathcal{O}\left(\sqrt{\frac{\log(1/\delta)}{n}}+\sqrt{\mathfrak{R}_{\text{SDR}}+\sqrt{\frac{\log(1/\delta)}{n}}}\right)\\
&&=\hat{\beta}(X,\hat{Y})\\ 
&&\ \ \ +\mathcal{O}\left(\sqrt{\frac{\log(1/\delta)}{n}}+\sqrt{\mathfrak{R}_{\text{SDR}}+\sqrt{\frac{\log(1/\delta)}{n}}}\right)\\
\end{eqnarray*}
Hence, with probability at least $1-2\delta$, we have that
\begin{eqnarray}\label{f1}
&&R_{\beta\ell,D_\rho}(\hat{f}_{n,\hat{\beta}})\nonumber\\
&&=R[D_\rho,\hat{f}_{n,\hat{\beta}},\beta(X,\hat{Y})\ell(\hat{f}_{n,\hat{\beta}}(X),\hat{Y})]\nonumber\\
&&=E_{(X,\hat{Y})\sim D_\rho}\left[\beta(X,\hat{Y})\ell(\hat{f}_{n,\hat{\beta}}(X),\hat{Y})\right]\nonumber\\
&&\leq E_{(X,\hat{Y})\sim D_\rho}\left[\left(\hat{\beta}(X,\hat{Y})\right.\right.\nonumber\\
&&\left.\left.\ \ \ +\mathcal{O}\left(\sqrt{\frac{\log(1/\delta)}{n}}+\sqrt{\mathfrak{R}_{\text{SDR}}+\sqrt{\frac{\log(1/\delta)}{n}}}\right)\right)\right.\nonumber\\ 
&&\left.\ \ \ \ell(\hat{f}_{n,\hat{\beta}}(X),\hat{Y})\right]\nonumber
\end{eqnarray}
\begin{eqnarray}\label{f1}
&&=R\left[D_\rho,\hat{f}_{n,\hat{\beta}},\hat{\beta}(X,\hat{Y})\ell(\hat{f}_{n,\hat{\beta}}(X),\hat{Y})\right]\nonumber\\
&&\ \ \ +\mathcal{O}\left(\sqrt{\frac{\log(1/\delta)}{n}}+\sqrt{\mathfrak{R}_{\text{SDR}}+\sqrt{\frac{\log(1/\delta)}{n}}}\right)\nonumber\\
&&=R_{\hat{\beta}\ell,D_\rho}(\hat{f}_{n,\hat{\beta}})\\ 
&&\ \ \ +\mathcal{O}\left(\sqrt{\frac{\log(1/\delta)}{n}}+\sqrt{\mathfrak{R}_{\text{SDR}}+\sqrt{\frac{\log(1/\delta)}{n}}}\right).\ \ \ \ \nonumber
\end{eqnarray}

Using the proof method of Proposition 1, with probability at least $1-\delta$, we have
\begin{eqnarray*}
&&R_{\hat{\beta}\ell,D_\rho}(\hat{f}_{n,\hat{\beta}})-R_{\hat{\beta}\ell,D_\rho}(f^*)\\
&&\leq2\sup_{f\in F}\left|E_{(X,\hat{Y})\sim D_\rho}\left[\hat{R}_{\hat{\beta}\ell,D_\rho}\right]-\hat{R}_{\hat{\beta}\ell,D_\rho}\right|\\
&&\leq 2\frac{1-\min_{(X,\hat{Y})}\frac{\rho_{-\hat{Y}}}{\hat{P}_{D_\rho}(\hat{Y}|X)}}{1-\rho_{-1}-\rho_{+1}}\mathfrak{R}(\ell\circ F)+2b\sqrt{\frac{\log(1/\delta)}{2n}}\\
&&\leq 2\frac{1-\min(\rho_{-1},\rho_{+1})}{1-\rho_{-1}-\rho_{+1}}\mathfrak{R}(\ell\circ F)+2b\sqrt{\frac{\log(1/\delta)}{2n}},
\end{eqnarray*}
or
\begin{eqnarray}\label{f2}
&&R_{\hat{\beta}\ell,D_\rho}(\hat{f}_{n,\hat{\beta}})-R_{\hat{\beta}\ell,D_\rho}(f^*)\nonumber\\
&&\leq\mathcal{O}\left( \mathfrak{R}(\ell\circ F)+\sqrt{\frac{\log(1/\delta)}{n}}\right).
\end{eqnarray}
Combining Equations (\ref{f1}) and (\ref{f2}), with probability at least $1-3\delta$, we have
\begin{eqnarray*}
R_{\beta\ell,D_\rho}(\hat{f}_{n,\hat{\beta}})-R_{\hat{\beta}\ell,D_\rho}(f^*)\ \ \ \ \ \ \ \ \ \ \ \ \ \ \ \ \ \ \ \ \ \ \ \ \ \ \ \ \ \ \ \ \ \ \ \ \ \\
\leq\mathcal{O}\left(\mathfrak{R}(\ell\circ F)+\sqrt{\frac{\log(1/\delta)}{n}}+\sqrt{\mathfrak{R}_{\text{SDR}}+\sqrt{\frac{\log(1/\delta)}{n}}}\right),
\end{eqnarray*}
which completes the proof. \hfill$\blacksquare$

\section{Proofs of the Assertions in Remark 2}

\subsection{Consistency of the Joint Estimation of the Noise Rate, Weight and Classifier}
We have considered the consistency of the joint estimation of the weight and classifier in Theorems 2 and 3. Those consistency results can be easily extended to the joint estimation of the noise rate, weight and classifier. In Lemma 3, we have proven that the kernel density estimation method can learn the target weight $\beta^* (X,\hat{Y})$ by employing a universal kernel. Thus, given sufficiently large data and the assumption in Theorem 4 that $\exists x_{-1},x_{+1}\in\mathcal{X},P_D(Y=+1|x_{-1})=P_D(Y=-1|x_{+1})=0$, Theorem 4 guarantees that we will learn the target noise rates. Theorem 2 therefore can be extended to provide a theoretical justification for the consistency of learning the optimal classifier in the hypothesis class with the estimated noise rates and weights. In Theorem 3, we have assumed that when employing the density ratio estimation method to learn the conditional distribution $P_{D_{\rho}}(\hat{Y}|X)$ and the hypothesis class is properly chosen, the corresponding approximation error is zero. The estimated noise rates therefore will converge to the target noise rates and the estimated  weights will approach to the target weights under the assumption in Theorem 4. Thus, Theorem 3 can be extended to the consistency of the joint estimation of the noise rate, weight and classifier as well. 

\subsection{Convergence rate of the Joint Estimation of the Noise Rate, Weight and Classifier}
We have discussed the convergence rate of the joint estimation of the weight and classifier in Proposition 2. In this subsection, we characterize a convergence rate for the joint estimation of the noise rate, weight and classifier. 

Let $\hat{P}_{D_\rho}(\hat{Y}|X)$ be an estimator for $P_{D_\rho}(\hat{Y}|X)$ using equations $(1)$, $(2)$ and $(3)$, and 
\begin{eqnarray}\label{noiserate}
&&\hat{\rho}_{-\hat{Y}}=\min_{X\in\{X_1,\ldots,X_n\}}\hat{P}_{D_\rho}(\hat{Y}|X).
\end{eqnarray}
be the estimators for $\rho_{\pm1}$.

Let defined the (learned) weight function comprised of the learned conditional distribution and learned noise rates as follows: 
\begin{eqnarray*}
&&\hat{\beta}(X,\hat{Y})=\frac{\hat{P}_{D_\rho}(\hat{Y}|X)-\hat{\rho}_{-\hat{Y}}}{(1-\hat{\rho}_{-1}-\hat{\rho}_{+1})\hat{P}_{D_\rho}(\hat{Y}|X)}.
\end{eqnarray*}
Note that the weight function is different from that defined in Theorem 2, where the noise rates are known and not estimated.

We also let
\begin{eqnarray*}
&&\hat{f}_{n,\hat{\beta}}=\min_{f\in F}\frac{1}{n}\sum_{i=1}^{n}\hat{\beta}(X_i,\hat{Y}_i)\ell(f(X_i),\hat{Y}_i)
\end{eqnarray*}
and
\begin{eqnarray*}
&&f^*=\min_{f\in F}R[D,f,\ell(f(X),{Y})].
\end{eqnarray*}
 
Then, the convergence rate for the joint estimation of the learned noise rate, weight and classifier is characterized as follows:
\begin{prop}\label{crate}
Under the settings of Theorem 3, if the Bregman divergence degenerates to square distance, for any $\delta>0$, with probability at least $1-9\delta$, the following holds:
\begin{eqnarray*}
&&R[D,\hat{f}_{n,\hat{\beta}},\ell(\hat{f}_{n,\hat{\beta}}(X),Y)]-R[D,f^*,\ell(f^*(X),Y)]\\
&&\leq{\scriptstyle\frac{\mathcal{O}\left(\sqrt{\frac{\log(1/\delta)}{n}}+\sqrt{\mathfrak{R}_{\text{SDR}}+\sqrt{\frac{\log(1/\delta)}{n}}}\right)}{\left(1-\hat{\rho}_{-1}-\hat{\rho}_{+1}-\mathcal{O}\left(\sqrt{\frac{\log(1/\delta)}{n}}+\sqrt{\mathfrak{R}_{\text{SDR}}+\sqrt{\frac{\log(1/\delta)}{n}}}\right)\right)^2}}\\
&&\ \ \ +\mathcal{O}\left(\mathfrak{R}(\ell\circ F)+\sqrt{\frac{\log(1/\delta)}{n}}\right),
\end{eqnarray*}
\end{prop}

\textbf{Proof of Proposition \ref{crate}.}
In the proof of Proposition 2, we have proven that with probability at least $1-2\delta$, we have
\begin{eqnarray}\label{upperb}
&&\left|P_{D_{\rho}}(\hat{Y}|X)- \hat{P}_{D_{\rho}}(\hat{Y}|X)\right|\\ 
&&\leq\mathcal{O}\left(\sqrt{\frac{\log(1/\delta)}{n}}+\sqrt{\mathfrak{R}_{\text{SDR}}+\sqrt{\frac{\log(1/\delta)}{n}}}\right).\nonumber
\end{eqnarray} 

We have also proven that with probability at least $1-2\delta$, we have
\begin{eqnarray*}
&&\frac{{P}_{D_\rho}(\hat{Y}|X)-\rho_{-\hat{Y}}}{(1-\rho_{-1}-\rho_{+1}){P}_{D_\rho}(\hat{Y}|X)}\\ 
&&\leq\frac{\hat{P}_{D_\rho}(\hat{Y}|X)-\rho_{-\hat{Y}}}{(1-\rho_{-1}-\rho_{+1})\hat{P}_{D_\rho}(\hat{Y}|X)}\\
&&\ \ \ +\mathcal{O}\left(\sqrt{\frac{\log(1/\delta)}{n}}+\sqrt{\mathfrak{R}_{\text{SDR}}+\sqrt{\frac{\log(1/\delta)}{n}}}\right).
\end{eqnarray*}

According to (\ref{noiserate}) and (\ref{upperb}), with probability at least $1-2\delta$, we have
\begin{eqnarray*}
|\hat{\rho}_{\pm1} -\rho_{\pm1}|\leq\mathcal{O}\left(\sqrt{\frac{\log(1/\delta)}{n}}+\sqrt{\mathfrak{R}_{\text{SDR}}+\sqrt{\frac{\log(1/\delta)}{n}}}\right).
\end{eqnarray*}
This is because, with probability at least $1-2\delta$, we have
\begin{eqnarray*}
&\hat{\rho}_{\pm1} -\rho_{\pm1}&=\min_{X\in\{X_1,\ldots,X_n\}}\hat{P}_{D_\rho}(\hat{Y}|X)-\min_{X\in\mathcal{X}}{P}_{D_\rho}(\hat{Y}|X)\\ 
&&\leq \max_{X\in\mathcal{X}}\hat{P}_{D_\rho}(\hat{Y}|X)-{P}_{D_\rho}(\hat{Y}|X)\\ 
&&\leq \mathcal{O}\left(\sqrt{\frac{\log(1/\delta)}{n}}+\sqrt{\mathfrak{R}_{\text{SDR}}+\sqrt{\frac{\log(1/\delta)}{n}}}\right)
\end{eqnarray*}
and
\begin{eqnarray*}
&\rho_{\pm1}-\hat{\rho}_{\pm1} &=\min_{X\in\mathcal{X}}{P}_{D_\rho}(\hat{Y}|X)-\min_{X\in\{X_1,\ldots,X_n\}}\hat{P}_{D_\rho}(\hat{Y}|X)\\ 
&&\leq \max_{X\in\mathcal{X}}{P}_{D_\rho}(\hat{Y}|X)-\hat{P}_{D_\rho}(\hat{Y}|X)\\ 
&&\leq \mathcal{O}\left(\sqrt{\frac{\log(1/\delta)}{n}}+\sqrt{\mathfrak{R}_{\text{SDR}}+\sqrt{\frac{\log(1/\delta)}{n}}}\right).
\end{eqnarray*}

Thus, with probability at least $1-4\delta$, it holds that
\begin{eqnarray}\label{r1}
&&\frac{\hat{P}_{D_\rho}(\hat{Y}|X)-\rho_{-\hat{Y}}}{(1-\rho_{-1}-\rho_{+1})\hat{P}_{D_\rho}(\hat{Y}|X)}\nonumber\\
&&\ \ \ +\mathcal{O}\left(\sqrt{\frac{\log(1/\delta)}{n}}+\sqrt{\mathfrak{R}_{\text{SDR}}+\sqrt{\frac{\log(1/\delta)}{n}}}\right)\nonumber\\
&&\leq\frac{\hat{P}_{D_\rho}(\hat{Y}|X)-\hat{\rho}_{-\hat{Y}}}{(1-\rho_{-1}-\rho_{+1})\hat{P}_{D_\rho}(\hat{Y}|X)}\\
&&\ \ \ +\mathcal{O}\left(\sqrt{\frac{\log(1/\delta)}{n}}+\sqrt{\mathfrak{R}_{\text{SDR}}+\sqrt{\frac{\log(1/\delta)}{n}}}\right).\nonumber
\end{eqnarray}

Since with probability at least $1-2\delta$, it holds that $\hat{\rho}_{\pm1}\geq \rho_{\pm1}-\mathcal{O}\left(\sqrt{\frac{\log(1/\delta)}{n}}+\sqrt{\mathfrak{R}_{\text{SDR}}+\sqrt{\frac{\log(1/\delta)}{n}}}\right)$, with probability at least $1-4\delta$, we have
\begin{eqnarray}\label{r2}
\frac{1}{1-\rho_{-1}-\rho_{+1}}\leq \frac{1}{1-\hat{\rho}_{-1}-\hat{\rho}_{+1}-\Delta(n)},
\end{eqnarray}
where $\Delta(n)\triangleq\mathcal{O}\left(\sqrt{\frac{\log(1/\delta)}{n}}+\sqrt{\mathfrak{R}_{\text{SDR}}+\sqrt{\frac{\log(1/\delta)}{n}}}\right)$.

We now prove that for any $a>0,b>0,a+b<1$, it holds that
\begin{eqnarray}\label{r3}
\frac{1}{1-a-b}\leq \frac{1}{1-a}+\frac{b}{(1-a-b)^2}.
\end{eqnarray}
This is because 
\begin{eqnarray*} 
&&\frac{1}{1-a-b}\leq \frac{1}{1-a}+\frac{b}{(1-a-b)^2}\\ 
&\Leftrightarrow& 1-a\leq 1-a-b +\frac{b(1-a)}{1-a-b}\\ 
&\Leftrightarrow& 1\leq \frac{1-a}{1-a-b}.
\end{eqnarray*}

Combining inequalities (\ref{r2}) and (\ref{r3}), with probability at least $1-4\delta$, we have
\begin{eqnarray}\label{r4}
&&\frac{1}{1-\rho_{-1}-\rho_{+1}}\nonumber\\ 
&&\leq \frac{1}{1-\hat{\rho}_{-1}-\hat{\rho}_{+1}-\Delta(n)}\\ 
&&\leq \frac{1}{1-\hat{\rho}_{-1}-\hat{\rho}_{+1}}+\frac{\Delta(n)}{(1-\hat{\rho}_{-1}-\hat{\rho}_{+1}-\Delta(n))^2}\nonumber
\end{eqnarray}

Combining inequalities (\ref{r1}) and (\ref{r4}), we have that with probability at least $1-8\delta$, the following holds
\begin{eqnarray}\label{r5}
&&{\beta}(X,\hat{Y})\nonumber\\ 
&&\leq\frac{\hat{P}_{D_\rho}(\hat{Y}|X)-\rho_{-\hat{Y}}}{(1-\rho_{-1}-\rho_{+1})\hat{P}_{D_\rho}(\hat{Y}|X)}\nonumber\\
&&\ \ \ +\mathcal{O}\left(\sqrt{\frac{\log(1/\delta)}{n}}+\sqrt{\mathfrak{R}_{\text{SDR}}+\sqrt{\frac{\log(1/\delta)}{n}}}\right)\nonumber\\
&&\leq\frac{\hat{P}_{D_\rho}(\hat{Y}|X)-\hat{\rho}_{-\hat{Y}}}{(1-\rho_{-1}-\rho_{+1})\hat{P}_{D_\rho}(\hat{Y}|X)}\nonumber\\
&&\ \ \ +\mathcal{O}\left(\sqrt{\frac{\log(1/\delta)}{n}}+\sqrt{\mathfrak{R}_{\text{SDR}}+\sqrt{\frac{\log(1/\delta)}{n}}}\right).\nonumber\\
&&\ \ \ (\text{According to $\hat{P}_{D_\rho}(\hat{Y}|X)-\hat{\rho}_{-\hat{Y}}\geq 0$ and (\ref{r4})})\nonumber\\
&&\leq\frac{\hat{P}_{D_\rho}(\hat{Y}|X)-\hat{\rho}_{-\hat{Y}}}{(1-\hat{\rho}_{-1}-\hat{\rho}_{+1})\hat{P}_{D_\rho}(\hat{Y}|X)}\nonumber\\
&&\ \ \ +{\scriptstyle\frac{\mathcal{O}\left(\sqrt{\frac{\log(1/\delta)}{n}}+\sqrt{\mathfrak{R}_{\text{SDR}}+\sqrt{\frac{\log(1/\delta)}{n}}}\right)}{\left(1-\hat{\rho}_{-1}-\hat{\rho}_{+1}-\mathcal{O}\left(\sqrt{\frac{\log(1/\delta)}{n}}+\sqrt{\mathfrak{R}_{\text{SDR}}+\sqrt{\frac{\log(1/\delta)}{n}}}\right)\right)^2}}\nonumber\\
&&\ \ \ +\mathcal{O}\left(\sqrt{\frac{\log(1/\delta)}{n}}+\sqrt{\mathfrak{R}_{\text{SDR}}+\sqrt{\frac{\log(1/\delta)}{n}}}\right).\nonumber\\
&&=\hat{\beta}(X,\hat{Y})\\
&&\ \ \ +{\scriptstyle\frac{\mathcal{O}\left(\sqrt{\frac{\log(1/\delta)}{n}}+\sqrt{\mathfrak{R}_{\text{SDR}}+\sqrt{\frac{\log(1/\delta)}{n}}}\right)}{\left(1-\hat{\rho}_{-1}-\hat{\rho}_{+1}-\mathcal{O}\left(\sqrt{\frac{\log(1/\delta)}{n}}+\sqrt{\mathfrak{R}_{\text{SDR}}+\sqrt{\frac{\log(1/\delta)}{n}}}\right)\right)^2}}\nonumber\\
&&\ \ \ +\mathcal{O}\left(\sqrt{\frac{\log(1/\delta)}{n}}+\sqrt{\mathfrak{R}_{\text{SDR}}+\sqrt{\frac{\log(1/\delta)}{n}}}\right).\nonumber
\end{eqnarray}

Hence, with probability at least $1-8\delta$, we have that
\begin{eqnarray}\label{f11}
&&R_{\beta\ell,D_\rho}(\hat{f}_{n,\hat{\beta}})\nonumber\\
&&=R[D_\rho,\hat{f}_{n,\hat{\beta}},\beta(X,\hat{Y})\ell(\hat{f}_{n,\hat{\beta}}(X),\hat{Y})]\nonumber\\
&&=E_{(X,\hat{Y})\sim D_\rho}\left[\beta(X,\hat{Y})\ell(\hat{f}_{n,\hat{\beta}}(X),\hat{Y})\right]\nonumber\\
&&\leq E_{(X,\hat{Y})\sim D_\rho}\left[\left(\hat{\beta}(X,\hat{Y})\right.\right.\nonumber\\
&&\left.\left.\ \ \ +{\scriptstyle\frac{\mathcal{O}\left(\sqrt{\frac{\log(1/\delta)}{n}}+\sqrt{\mathfrak{R}_{\text{SDR}}+\sqrt{\frac{\log(1/\delta)}{n}}}\right)}{\left(1-\hat{\rho}_{-1}-\hat{\rho}_{+1}-\mathcal{O}\left(\sqrt{\frac{\log(1/\delta)}{n}}+\sqrt{\mathfrak{R}_{\text{SDR}}+\sqrt{\frac{\log(1/\delta)}{n}}}\right)\right)^2}}\right.\right.\nonumber\\
&&\left.\left.\ \ \ +\mathcal{O}\left(\sqrt{\frac{\log(1/\delta)}{n}}+\sqrt{\mathfrak{R}_{\text{SDR}}+\sqrt{\frac{\log(1/\delta)}{n}}}\right)\right)\right.\nonumber\\ 
&&\left.\ \ \ \ell(\hat{f}_{n,\hat{\beta}}(X),\hat{Y})\right]\nonumber\\
&&=R\left[D_\rho,\hat{f}_{n,\hat{\beta}},\hat{\beta}(X,\hat{Y})\ell(\hat{f}_{n,\hat{\beta}}(X),\hat{Y})\right]\nonumber\\
&&\ \ \ +{\scriptstyle\frac{\mathcal{O}\left(\sqrt{\frac{\log(1/\delta)}{n}}+\sqrt{\mathfrak{R}_{\text{SDR}}+\sqrt{\frac{\log(1/\delta)}{n}}}\right)}{\left(1-\hat{\rho}_{-1}-\hat{\rho}_{+1}-\mathcal{O}\left(\sqrt{\frac{\log(1/\delta)}{n}}+\sqrt{\mathfrak{R}_{\text{SDR}}+\sqrt{\frac{\log(1/\delta)}{n}}}\right)\right)^2}}\nonumber\\
&&\ \ \ +\mathcal{O}\left(\sqrt{\frac{\log(1/\delta)}{n}}+\sqrt{\mathfrak{R}_{\text{SDR}}+\sqrt{\frac{\log(1/\delta)}{n}}}\right)\nonumber\\
&&=R_{\hat{\beta}\ell,D_\rho}(\hat{f}_{n,\hat{\beta}})\nonumber\\ 
&&\ \ \ +{\scriptstyle\frac{\mathcal{O}\left(\sqrt{\frac{\log(1/\delta)}{n}}+\sqrt{\mathfrak{R}_{\text{SDR}}+\sqrt{\frac{\log(1/\delta)}{n}}}\right)}{\left(1-\hat{\rho}_{-1}-\hat{\rho}_{+1}-\mathcal{O}\left(\sqrt{\frac{\log(1/\delta)}{n}}+\sqrt{\mathfrak{R}_{\text{SDR}}+\sqrt{\frac{\log(1/\delta)}{n}}}\right)\right)^2}}\nonumber\\
&&\ \ \ +\mathcal{O}\left(\sqrt{\frac{\log(1/\delta)}{n}}+\sqrt{\mathfrak{R}_{\text{SDR}}+\sqrt{\frac{\log(1/\delta)}{n}}}\right).\ \ \ \ 
\end{eqnarray}

Using the proof method of Proposition 1, with probability at least $1-\delta$, we have
\begin{eqnarray*}
&&R_{\hat{\beta}\ell,D_\rho}(\hat{f}_{n,\hat{\beta}})-R_{\hat{\beta}\ell,D_\rho}(f^*)\\
&&\leq2\sup_{f\in F}\left|E_{(X,\hat{Y})\sim D_\rho}\left[\hat{R}_{\hat{\beta}\ell,D_\rho}\right]-\hat{R}_{\hat{\beta}\ell,D_\rho}\right|\\
&&\leq 2\frac{1-\min_{(X,\hat{Y})}\frac{\hat{\rho}_{-\hat{Y}}}{\hat{P}_{D_\rho}(\hat{Y}|X)}}{1-\hat{\rho}_{-1}-\hat{\rho}_{+1}}\mathfrak{R}(\ell\circ F)+2b\sqrt{\frac{\log(1/\delta)}{2n}}\\
&&\leq 2\frac{1-\min(\hat{\rho}_{-1},\hat{\rho}_{+1})}{1-\hat{\rho}_{-1}-\hat{\rho}_{+1}}\mathfrak{R}(\ell\circ F)+2b\sqrt{\frac{\log(1/\delta)}{2n}},
\end{eqnarray*}
or
\begin{eqnarray}\label{f12}
&&R_{\hat{\beta}\ell,D_\rho}(\hat{f}_{n,\hat{\beta}})-R_{\hat{\beta}\ell,D_\rho}(f^*)\nonumber\\
&&\leq\mathcal{O}\left( \mathfrak{R}(\ell\circ F)+\sqrt{\frac{\log(1/\delta)}{n}}\right).
\end{eqnarray}
Combining Equations (\ref{f11}) and (\ref{f12}), with probability at least $1-9\delta$, we have
\begin{eqnarray*}
&&R_{\beta\ell,D_\rho}(\hat{f}_{n,\hat{\beta}})-R_{\hat{\beta}\ell,D_\rho}(f^*)\\
&&\leq{\scriptstyle\frac{\mathcal{O}\left(\sqrt{\frac{\log(1/\delta)}{n}}+\sqrt{\mathfrak{R}_{\text{SDR}}+\sqrt{\frac{\log(1/\delta)}{n}}}\right)}{\left(1-\hat{\rho}_{-1}-\hat{\rho}_{+1}-\mathcal{O}\left(\sqrt{\frac{\log(1/\delta)}{n}}+\sqrt{\mathfrak{R}_{\text{SDR}}+\sqrt{\frac{\log(1/\delta)}{n}}}\right)\right)^2}}\\
&&\ \ \ +\mathcal{O}\left(\mathfrak{R}(\ell\circ F)+\sqrt{\frac{\log(1/\delta)}{n}}\right),
\end{eqnarray*}
which completes the proof. \hfill$\blacksquare$

\end{appendices}
\bibliographystyle{ieeetr}

\end{document}